\documentclass[]{fairmeta}
\setcitestyle{sort}

\newcommand{\suggestion}[2]{
    \begin{tcolorbox}[
        colback=blue!5,
        colframe=blue!70!black,
        arc=5pt,
        boxsep=5pt,
        left=2pt,
        right=2pt,
        top=2pt,
        bottom=2pt,
        boxrule=0.8pt,
        drop shadow=gray!50!white,
        enhanced jigsaw,
        before skip=12pt plus 2pt minus 2pt, 
        after skip=12pt plus 2pt minus 2pt,  
    ]
        \phantomsection\label{suggestion:#1}%
        \noindent\textbf{\textit{Suggestion #1:}} #2
    \end{tcolorbox}
}

\newcommand{\suggref}[2]{\hyperref[suggestion:#1]{\textcolor{blue!50!black}{\textit{#2}}}}

\usepackage[dvipsnames]{xcolor}
\iftrue   
    \newcommand{\displaytodo}[1]{#1}
\else
    \newcommand{\displaytodo}[1]{}
\fi

\usepackage{makecell}
\usepackage{epsfig}
\usepackage{amsmath}
\usepackage{amssymb}
\usepackage{fontawesome5}
\usepackage{algorithmic}
\usepackage{colortbl} 
\usepackage{arydshln}
\usepackage{tikz}
\usetikzlibrary{tikzmark}
\usepackage{siunitx} 
\usepackage{mdframed}
\usepackage{tablefootnote} 
\usepackage{xspace}
\newcolumntype{g}{>{\columncolor{gray!30}}c}
\usepackage{wrapfig}
\definecolor{citecolor}{HTML}{0071bc}

\definecolor{fairblue}{HTML}{0064E0}
\definecolor{nyupurple}{HTML}{57068C}

\usepackage{fdsymbol}

\newcommand{\fairmark}{\textcolor{fairblue}{1}}
\newcommand{\nyumark}{\textcolor{nyupurple}{2}}

\tcbuselibrary{skins}

\definecolor{userbg}{RGB}{245, 245, 245}
\definecolor{userborder}{RGB}{210, 229, 255}
\definecolor{userfont}{RGB}{0, 0, 0}

\tcbset{
  enhanced,
  boxrule=1pt,
  colback=userbg,
  colframe=userborder,
  fonttitle=\bfseries,
  coltitle=userfont,
  boxsep=5pt,
  arc=5pt,
  outer arc=5pt,
  left=10pt,
  right=10pt,
  top=2pt,
  bottom=2pt,
  attach boxed title to top left={yshift=-0.25\baselineskip-1mm,xshift=2mm},
  boxed title style={
    colback=userborder,
    sharp corners,
    rounded corners=northeast,
    arc=3pt,
    outer arc=3pt,
    boxrule=0pt,
    boxsep=2pt,
  },
}
\usepackage{pgf}
\usepackage{adjustbox}
\usepackage[shortlabels,inline]{enumitem} 
\usepackage{array}
\definecolor{listcolor}{RGB}{50,120,230}

\definecolor{methodgreen}{RGB}{45, 106, 79} 

\newcommand{\mix}[1]{{\fontfamily{lmss}\selectfont #1}}

\usepackage{mathtools}
\usepackage{amsthm}

\titlespacing*{\paragraph} {0pt}{1.25ex plus 1ex minus .2ex}{0.75em}

\definecolor{DiTblue1}{HTML}{AEC6FE}
\definecolor{DiTblue2}{HTML}{9AB8FE}
\definecolor{DiTblue3}{HTML}{86AAFE}
\definecolor{DiTblue4}{HTML}{729CFE}
\definecolor{DiTblue5}{HTML}{5D8DFD}

\definecolor{DiTgreen1}{HTML}{B3CDA2}
\definecolor{DiTgreen2}{HTML}{A8C695}
\definecolor{DiTgreen3}{HTML}{9EBF88}
\definecolor{DiTgreen4}{HTML}{93B87A}
\definecolor{DiTgreen5}{HTML}{88B06D}

\definecolor{DiTyellow1}{HTML}{F9CB8A}
\definecolor{DiTyellow2}{HTML}{F8C277}
\definecolor{DiTyellow3}{HTML}{F7BA64}
\definecolor{DiTyellow4}{HTML}{F6B150}
\definecolor{DiTyellow5}{HTML}{F5A83D}

\definecolor{DiTred1}{HTML}{FBA09D}
\definecolor{DiTred2}{HTML}{FA8480}
\definecolor{DiTred3}{HTML}{F86762}
\definecolor{DiTred4}{HTML}{F65550}
\definecolor{DiTred5}{HTML}{F5433D}
\definecolor{EncoderSigLIP 2}{HTML}{729CFE}  
\definecolor{EncoderWebSSL}{HTML}{A8C695}   
\definecolor{EncoderDiNOv2}{HTML}{729CFE}   
\definecolor{EncoderSDVAE}{HTML}{F86762}    
\definecolor{EncoderFLUX}{HTML}{F65550}     
\definecolor{EncoderRawPixel}{HTML}{F7BA64} 

\definecolor{DataDrivenPurple}{HTML}{897bc1}
\definecolor{ExtendDiTPurple}{HTML}{D4B6FC}
\newcommand{\texttok}{\textcolor{DiTblue5}{\texttt{T}}\xspace}
\newcommand{\imtok}{\textcolor{DiTred4}{\texttt{I}}\xspace}

\newcommand{\Language}{\textcolor{DiTblue4}{\textbf{Language}}\xspace}
\newcommand{\Vision}{\textcolor{DiTred4}{\textbf{Vision}}\xspace}

\theoremstyle{plain}

\theoremstyle{definition}

\theoremstyle{remark}

\title{Beyond Language Modeling:\\An Exploration of Multimodal Pretraining}

\author[\fairmark, \nyumark, *]{Shengbang~Tong}
\author[\fairmark, *]{David~Fan}
\author[\fairmark, *]{John~Nguyen}
\author[\fairmark, \nyumark]{\\[0.25em] Ellis~Brown}
\author[\fairmark, \nyumark]{Gaoyue~Zhou}
\author[\fairmark]{Shengyi~Qian}
\author[\nyumark]{Boyang~Zheng}
\author[\fairmark]{Théophane~Vallaeys}
\author[\fairmark]{Junlin~Han}
\author[\fairmark, \nyumark]{Rob~Fergus}
\author[\fairmark]{Naila~Murray}
\author[\fairmark]{Marjan~Ghazvininejad}
\author[\fairmark]{Mike~Lewis}
\author[\fairmark]{Nicolas~Ballas}
\author[\fairmark]{Amir~Bar}
\author[\fairmark]{Michael~Rabbat}
\author[\fairmark]{Jakob~Verbeek}
\author[\fairmark, \dagger]{\\[0.25em] Luke~Zettlemoyer}
\author[\fairmark, \dagger]{Koustuv~Sinha}
\author[\nyumark, \dagger]{Yann~LeCun}
\author[\nyumark, \dagger]{Saining~Xie}

\affiliation[\fairmark]{\textcolor{fairblue}{FAIR, Meta}}
\affiliation[\nyumark]{\textcolor{nyupurple}{New York University}}
\contribution[*]{equal contribution}
\contribution[\dagger]{equal advising}

\abstract{
The visual world offers a critical axis for advancing foundation models beyond language. Despite growing interest in this direction, the design space for native multimodal models remains opaque. We provide empirical clarity through controlled, from-scratch pretraining experiments, isolating the factors that govern multimodal pretraining without interference from language pretraining. We adopt the Transfusion framework, using next-token prediction for language and diffusion for vision, to train on diverse data including text, video, image-text pairs, and even action-conditioned video.
Our experiments yield four key insights:
(i) Representation Autoencoder (RAE) provides an optimal unified visual representation by excelling at both visual understanding and generation;
(ii) visual and language data are complementary and yield synergy for downstream capabilities;
(iii) unified multimodal pretraining leads naturally to world modeling, with capabilities emerging from general training; 
and (iv) Mixture-of-Experts (MoE) enables efficient and effective multimodal scaling while naturally inducing modality specialization.
Through IsoFLOP analysis, we compute scaling laws for both modalities and uncover a scaling asymmetry: vision is significantly more data-hungry than language. We demonstrate that the MoE architecture harmonizes this scaling asymmetry by providing the high model capacity required by language while accommodating the data-intensive nature of vision, paving the way for truly unified multimodal models.
}

\date{March 3, 2026}

\project{\href{https://beyond-llms.github.io/}{\sffamily \fontsize{8.8pt}{11pt}\selectfont \texttt{https://beyond-llms.github.io/}}}

\begin{document}

\maketitle

\vspace{3em}
\noindent
\begin{tcolorbox}[
    colback=blue!3,
    colframe=blue!60!black,
    arc=4pt,
    boxrule=0.5pt,
    boxsep=3pt,
    left=6pt, right=6pt, top=4pt, bottom=3pt,
    enhanced jigsaw,
    drop shadow=gray!40!white,
]
{\noindent\sffamily\bfseries Suggestions at a Glance}\\[-2pt]
{\color{blue!30!black}\hrule height 0.3pt}\vspace{4pt}
\begin{enumerate}[leftmargin=1.5em, itemsep=6pt, parsep=0pt, topsep=0pt, label={\small\textbf{S\arabic*.}}]
    \vspace{6pt}
    \item \textbf{Unify visual representations}: RAE bridges visual understanding and generation.
    \hfill \suggref{1}{\S\ref*{sec:ablation_visual_representation}}
    \item \textbf{Embrace diverse data}: multimodal co-training yields cross-modal synergy. 
    \hfill \suggref{2}{\S\ref*{sec:data_composition}}
    \item \textbf{Unlock world modeling}: capabilities naturally emerge from unified multimodal pretraining.
    \hfill \suggref{3}{\S\ref*{sec:world_modeling}}
    \item \textbf{Use MoE}: MoE scales multimodal training efficiently and harmonizes capacity across modalities.
    \hfill \suggref{4}{\S\ref*{sec:arch_design},\ref*{sec:scaling}}
\end{enumerate}
\end{tcolorbox}

\clearpage
{
  \hypersetup{linkcolor=black}
  \tableofcontents
}
\clearpage

\section{Introduction}

The foundation model era has been defined largely by the success of language pretraining~\citep{OpenAI2022ChatGPT, Gemini}. By scaling autoregressive models on trillions of text tokens, we have created systems with remarkable reasoning capabilities. Yet, fundamentally, text is a human abstraction---a lossy compression of reality. To borrow the allegory of Plato’s Cave~\citep{plato_republic}, language models have mastered the description of shadows on the wall without ever seeing the objects casting them. They capture symbols well but miss the high-fidelity physics, geometry, and causality of the physical world.

Beyond this philosophical limitation lies a hard, pragmatic ceiling: high-quality text data is finite and approaching exhaustion~\citep{sutskever2025}. In contrast, the visual world possesses an endless stream of signal ``outside the cave'', capturing the raw dynamics of reality that language misses. As a result, the path forward requires moving beyond the shadows to model the source directly. We turn to unified multimodal pretraining, treating the visual signal not as an auxiliary input, but as a first-class citizen alongside language. 

The scientific landscape of unified multimodal pretraining remains largely opaque. While recent efforts~\citep{emu3, zhou2024transfusion,cui2025emu3} have begun to move beyond language-only pretraining, the design space is rife with confounding variables. Rather than jointly learning from vision and language from scratch, most current methodologies~\citep{mogao, deng2025bagel} rely on initialization from pretrained language models~\citep{grattafiori2024llama3,qwen3}. This paradigm prioritizes preserving existing language capabilities while adapting the model to become multimodal. Moreover, the knowledge already embedded in these pretrained backbones confounds any conclusions drawn about the multimodal training itself, making it difficult to disentangle what is learned from unified training versus what is inherited from language pretraining. Consequently, the fundamental dynamics and scaling relationships between vision and language remain poorly understood.

In this work, we seek to provide empirical clarity to this landscape. We focus squarely on pretraining, as a model's core capabilities are largely acquired during this phase~\citep{zhou2024lima}. We train a single model from scratch using the Transfusion framework, with next-token prediction for language and diffusion for vision, on diverse data including text, video, image-text pairs, and action-conditioned video. We conduct controlled experiments to isolate key variables and evaluate across a comprehensive spectrum of tasks, ranging from language evaluation and visual understanding/generation, to planning capabilities in world modeling.

Specifically, we investigate the following axes:

\begin{itemize}
    \item \textbf{Visual Representation:} We evaluate a broad spectrum of visual representations, ranging from Variational Autoencoders (VAE) and semantic representations to raw pixels. We identify Representation Autoencoders (RAE) as the optimal representation. (\Cref{sec:ablation_visual_representation}) 
    
\item \textbf{Data:} We study a wide range of data mixtures, from pure text and video to paired image-text and action-conditioned video. We find minimal interference between modalities and even observe positive synergy in certain cases. (\Cref{sec:data_composition})

\item \textbf{World Modeling:} We extend our evaluation to the Navigation World Model (NWM) setting, formatting actions directly as text tokens. We demonstrate that physical prediction capabilities emerge primarily from general multimodal pretraining (e.g. video) rather than domain-specific data. (\Cref{sec:world_modeling})

\item \textbf{Architecture:} We study design choices for Mixture-of-Experts (MoE) in this unified multimodal setting and observe natural formation of modality separation and unification.  (\Cref{sec:arch_design})

\item \textbf{Scaling Properties:} We conduct IsoFLOP experiments to derive scaling laws for vision and language during unified pretraining. We uncover a scaling asymmetry where vision is significantly more data-hungry than language, and find that MoE architectures effectively bridge this gap. (\Cref{sec:scaling})
\end{itemize}

\begin{figure}[t]
  \centering
  \includegraphics[width=\textwidth]{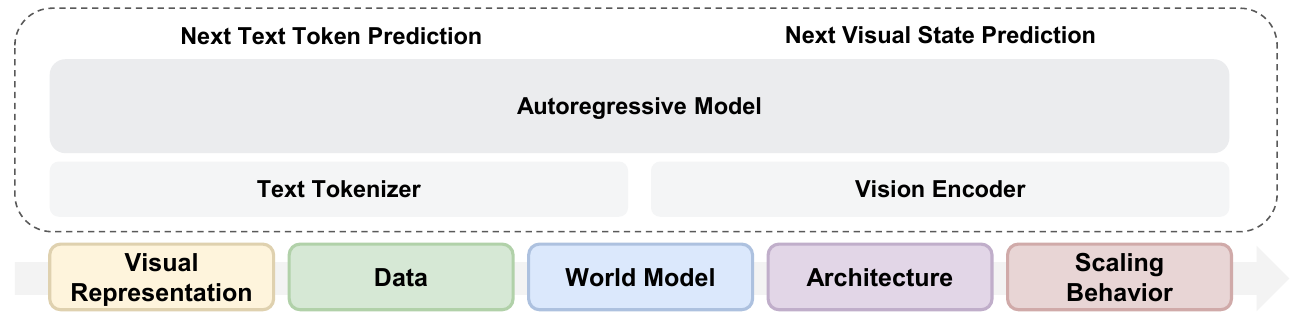}
  \caption{
    \textbf{Overview of our study.}
    \textbf{Top:} High-level model architecture. We train a single autoregressive model with next-token prediction for text and next-visual state prediction with flow matching.
    \textbf{Bottom:} We study five axes: visual representations, data, world modeling, architecture, and scaling.
  }\label{fig:pipeline}
\end{figure}

\section{Experiment Setup}
\label{sec:method}

Our goal is to study the science and design choices around unified multimodal pretraining.
Across experiments, we keep the compute budget and training hyperparameters controlled within each ablation family.
Unless otherwise specified, we evaluate at the end of pretraining (without any instruction tuning). 
We provide an overview of our training and inference protocols here, and more details in~\Cref{sec:extended_methodology}.

\subsection{Training and Inference}
\label{sec:method_training_and_inference}

\paragraph{Language modeling.}
Given a text sequence $(x_1, \dots, x_n)$, we minimize standard autoregressive cross-entropy:
\begin{equation}
    \mathcal{L}_{\text{LM}} = -\sum_{i=1}^{n}\log p_\theta(x_i \mid x_{<i}).
\end{equation}

\paragraph{Image-wise flow matching for visual tokens.}
Let $z_0$ denote the clean latents for an image or a video frame (flattened into a sequence), and let $\epsilon\sim\mathcal{N}(0,I)$.
Following flow matching formulations~\citep{fm,rf}, we sample $t\sim\mathcal{U}[0,1]$ and construct an interpolated latent
$z_t = (1-t)\,\epsilon + t\,z_0$.
The model predicts a velocity field $v_\theta(z_t, t, \text{context})$ and we minimize the squared error
\begin{equation}
\mathcal{L}_{\text{flow}} \;=\; \mathbb{E}_{t,z_0,\epsilon}\left[\,\lVert v_\theta(z_t,t,\cdot) - (z_0-\epsilon)\rVert_2^2\,\right].
\end{equation}
To facilitate diffusing through high-dimensional visual representations, we shift the noise schedule towards the noisier end of the spectrum~\citep{SD3}, following RAE~\citep{zheng2025diffusion, scale-rae-2026}.
For each image/frame, a single independent $t$ is sampled and applied to all tokens in that image. This independent noise injection aligns the training dynamics with Diffusion Forcing~\citep{chen2024diffusion}.
In \Cref{sec:ablation_visual_representation,sec:moe}, we additionally explore alternative parameterizations such as predicting $z_0$ directly (``$x$-pred'')~\citep{li2025jit}.

\begin{figure}[t]
  \centering
  \includegraphics[width=\textwidth]{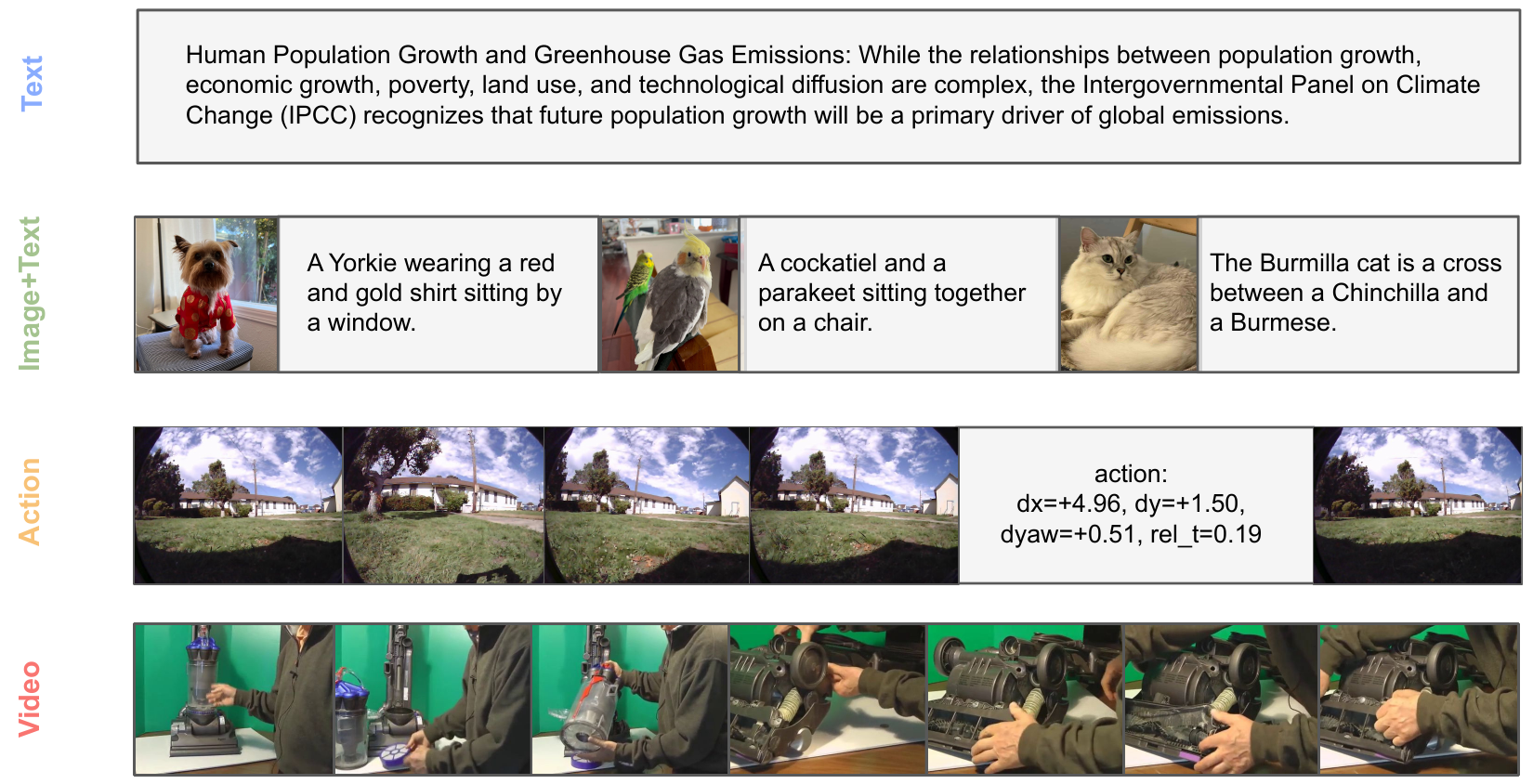}
  \caption{
    \textbf{Examples of training data.} 
    We train on diverse data including text, videos, image-text, and action-conditioned video.
  }\label{fig:data sources}
\end{figure}

\paragraph{Mixture training.}
Each batch contains a mixture of examples drawn from several sources (text-only, video-only, paired image--text, and action-conditioned data).
We optimize a weighted combination of objectives:
\begin{equation}
\mathcal{L} = \lambda_{\text{LM}}\mathcal{L}_{\text{LM}} + \lambda_{\text{flow}}\mathcal{L}_{\text{flow}},
\end{equation}
with weights chosen to stabilize joint training.
Unless otherwise specified, we use $\lambda_{\text{LM}}{=}1.0$ and $\lambda_{\text{flow}}{=}3.0$.

\paragraph{Causal vision and language masking.}

Training uses a hybrid masking strategy implemented via FlexAttention~\citep{dong2024flex}. Language modeling uses a standard causal mask. In contrast, visual data uses a block-wise causal mask where each video frame (or static image) is a separate block. Tokens from the same frame attend bidirectionally, and also to previous tokens. This allows us to operate on any combination of sequential data from any modality.

\paragraph{Inference.} Text generation follows standard autoregressive next-token prediction. For visual generation, the model denoises using a 25-step Euler sampler. The denoised latent is finally mapped back to pixel space using a pretrained decoder. For the VAE encoders, we use the off-the-shelf pretrained decoders from SD-VAE~\citep{LDM} and FLUX.1~\citep{flux}. For SigLIP 2 So400m, DINOv2-L, and WebSSL-L, we use the decoders from RAE~\citep{zheng2025diffusion, scale-rae-2026}. During training, we randomly drop the conditioning 10\% of the time to enable classifier-free guidance at inference time~\citep{ho2022classifier}. We use a fixed CFG of 3.0 to run all image generation evaluations and do not tune this value.

\subsection{Model and Tokenization}
\paragraph{Unified decoder-only backbone.}

\begin{wrapfigure}{r}{0.4\textwidth} 
  \centering
  \vspace{-10pt} 
  \includegraphics[width=\linewidth]{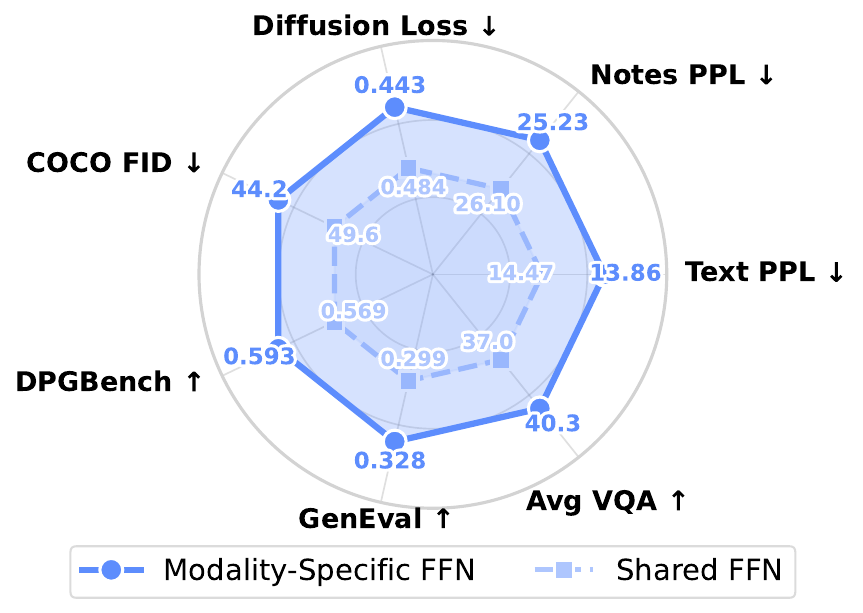}
  \caption{\textbf{Shared $\rightarrow$ Modality-specific FFN.} Modality separation improves multimodal learning over shared FFN. We explore MoE design in \Cref{sec:moe}.}
  \label{fig:ffn_shared_vs_specific}
  \vspace{-10pt} 
\end{wrapfigure}
We use a decoder-only Transformer backbone~\citep{vaswani2017attention} that is similar to Transfusion~\citep{zhou2024transfusion}, and train \emph{from scratch} on a mixture of vision and language data. 
Unlike Transfusion, we use simple linear projection layers instead of a U-Net~\citep{unet}. This design choice is motivated by the fact that we do not adjust the number of visual tokens for most encoders, allowing for a more direct mapping into the transformer's latent space. Within each self-attention block, we use modality-specific FFNs (one for text and one for vision tokens) instead of shared FFNs by default. We verify this design choice in~\Cref{fig:ffn_shared_vs_specific}: modality-specific FFNs unanimously improves performance---reducing text perplexity while improving image generation and VQA, consistent with prior findings~\citep{lin2024moma}. See~\Cref{sec:ablation_ffn_shared_vs_specific} for details. Our default model has 2.3B total parameters and 1.5B are activated per token due to the modality-specific FFNs; for comparison, a standard dense model with shared FFNs would also contain 1.5B parameters. In \Cref{sec:moe}, we will go beyond modality-specific FFN layers and explore learnable modality-wise FFN routing via MoEs.

\paragraph{Text tokens.}
Text is tokenized with a standard BPE tokenizer from LLaMA-3~\citep{grattafiori2024llama3} and trained with next-token prediction. We denote text tokens as \texttok.

\paragraph{Visual tokens.}
An image (or a video frame) is mapped to latent tokens using a \textit{frozen} visual encoder. Our architecture is encoder-agnostic and can accept any visual input. In \Cref{sec:ablation_visual_representation}, we explore a range of vision encoders ranging from semantic encoders (e.g.,\ SigLIP 2~\citep{tschannen2025siglip}, DINOv2~\citep{Dinov2}) to VAEs~\citep{VAE} to raw pixels. By default, we use SigLIP 2 So400M, unless otherwise specified. We denote images tokens as \imtok. In this work, because we use image encoders, videos are encoded frame-by-frame. All frames are pre-processed to 224$\times$224 pixels.

\subsection{Data Sources}
\label{sec:data_sources}

We use four broad categories of training data:
(i) large-scale web text (DCLM~\citep{li2024datacomp}),
(ii) raw videos (YouTube and curated video datasets~\citep{kay2017kinetics, goyal2017something, zellers2022merlot}) at 1 FPS,
(iii) paired image--text data (MetaCLIP~\citep{xu2023demystifying} and in-house Shutterstock),
and (iv) action-conditioned navigation trajectories formatted as \imtok + \texttok $\rightarrow$ \imtok (predicting a future visual state conditioned on a language action) taken from NWM~\citep{bar2025navigation} and annotated video data (see \Cref{appendix:video_action_annotation} for details). We showcase some examples in \Cref{fig:data sources} and present more details in \Cref{appendix: data sources}.

\subsection{Evaluation}
\label{sec:evaluation}
We report:
(i) \textbf{text perplexity} (PPL) on the held-out DCLM~\citep{li2024datacomp} validation set and in-house ``Notes'' text corpus for more OOD evaluation,
(ii) \textbf{diffusion loss} on the held-out CC12M~\citep{changpinyo2021conceptual} validation set, (iii) \textbf{image generation score} on DPGBench~\citep{hu2024ella} and GenEval~\citep{ghosh2023geneval},
and (iv) \textbf{average VQA accuracy} on the 16 Cambrian evaluation benchmarks~\citep{tong2024cambrian}. All VQA results are reported with 1 epoch of finetuning on the Cambrian-7M dataset.

\section{Visual Representations for Unified Multimodal Pretraining}
\label{sec:ablation_visual_representation}

\begin{figure}[t]
  \centering
  \includegraphics[width=\textwidth]{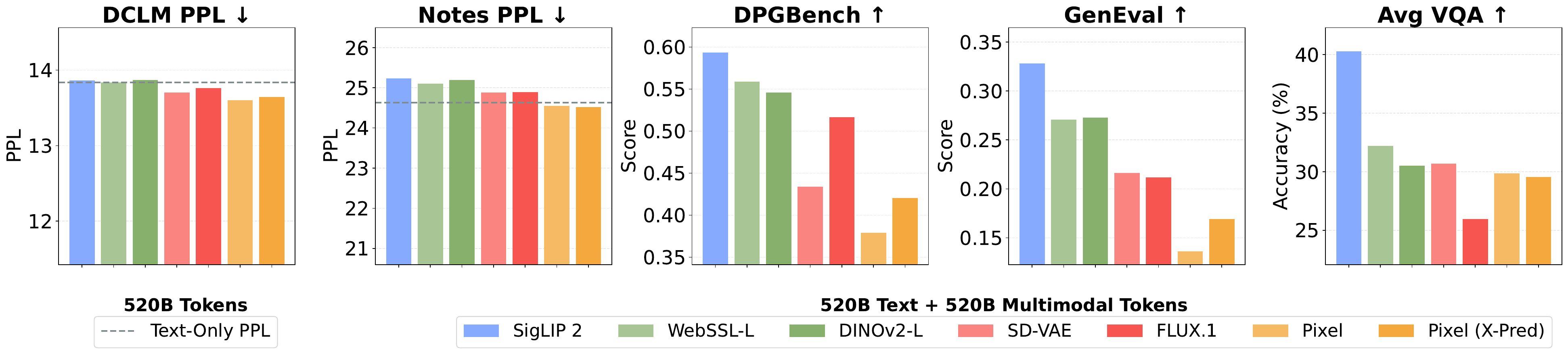}
  \caption{
    \textbf{RAE outperforms VAEs for both generation and understanding.}
    RAE (SigLIP 2) achieves the best performance on DPGBench, GenEval, and VQA while maintaining text perplexity comparable to the text-only baseline.}
  \label{fig:encoder_ablation}
\end{figure}

It is well understood that semantic encoders perform better for understanding tasks such as VQA~\citep{tong2024cambrian}. Yet, it has been widely assumed that VAE is necessary for visual generation~\citep{pgv3, ImprovDiffus}. Consequently, approaches like Janus~\citep{janus, ma2025janusflow}, BAGEL~\citep{deng2025bagel} and others~\citep{ai2025ming,fan2025unified} use dual representations for generation and understanding, such as VAE and SigLIP 2 respectively. However, this dichotomy significantly complicates model design and adds overhead to both model training and inference.

In contrast, recent studies on Representation Autoencoders (RAE)~\citep{zheng2025diffusion, scale-rae-2026} demonstrate that diffusion models can operate effectively within high-dimensional latent spaces. Leveraging these findings, we simplify our design by using a \textit{single} visual encoder for both understanding and generation.

We study three families of encoders. First, we study VAEs from Stable Diffusion (SD-VAE)~\citep{LDM} and FLUX.1~\citep{labs2025flux1kontextflowmatching}. Second, we study semantic encoders including both language-supervised encoders such as SigLIP 2 So400m~\citep{tschannen2025siglip}, and self-supervised encoders such as DINOv2-L~\citep{Dinov2} and WebSSL-L~\citep{fan2025scaling}. We use the off-the-shelf decoder from RAE to decode latents into pixels.  Finally, we study using raw pixels as direct inputs, where we patchify each image into tokens that each contain the flattened RGB values of a 14$\times$14  patch. Results are shown in \Cref{fig:encoder_ablation}.

\paragraph{Text performance.} We observe that all visual representations achieve similar (and sometimes even slightly better) text perplexity compared to the text-only baseline, with raw pixels performing the best. However, the marginal difference indicates that multimodal pretraining does not significantly affect the model's language capabilities compared to training on text alone, regardless of the visual representation.

\paragraph{Visual generation and understanding.} Semantic encoders consistently outperform VAE-based encoders on both visual understanding and generation. For example, SigLIP 2 not only outperforms FLUX.1 on VQA, but also on image generation benchmarks such as DPGBench and GenEval. This echoes findings from RAE~\citep{zheng2025diffusion, scale-rae-2026} that high-dimensional visual representations are as effective as, if not more effective than, low-dimensional VAE latents for generation. It shows that a single encoder suffices for both visual understanding and generation. Thus, we adopt SigLIP 2 as the default vision encoder.

We also evaluate training directly on raw pixels, with optionally the $x$-prediction parameterization from JiT~\citep{li2025jit}. We observe that while using raw pixels underperforms semantic encoders in generation quality, the performance gap in VQA accuracy is comparatively narrow. Given this competitive understanding performance, we believe that learning directly from raw pixels remains a promising direction for future research, particularly with more compute and scale.

\suggestion{1}{A single RAE-based encoder (e.g. SigLIP 2) simplifies the architecture by excelling at both visual understanding and generation, while preserving text performance.}

\section{Understanding the Impact of Data}
\label{sec:data_composition}

We study the impact of pretraining data composition. Our unified framework (\Cref{sec:method}) is capable of modeling any combination of multimodal data, such as text, video, image-text pairs (I/T), and action-conditioned video. We first compare text-only pretraining against multimodal pretraining (\Cref{sec:ablation_pretrain_data}), then dissect the role of text distribution in I/T data (\Cref{sec:ablation_image_text_distribution}), and finally demonstrate positive transfer from multimodal co-training (\Cref{sec:data_synergy}).

\subsection{Pretraining Data Composition}
\label{sec:ablation_pretrain_data}

\begin{figure*}[t]
    \centering
    \begin{minipage}[b]{0.49\textwidth}
        \centering
        \includegraphics[width=\linewidth]{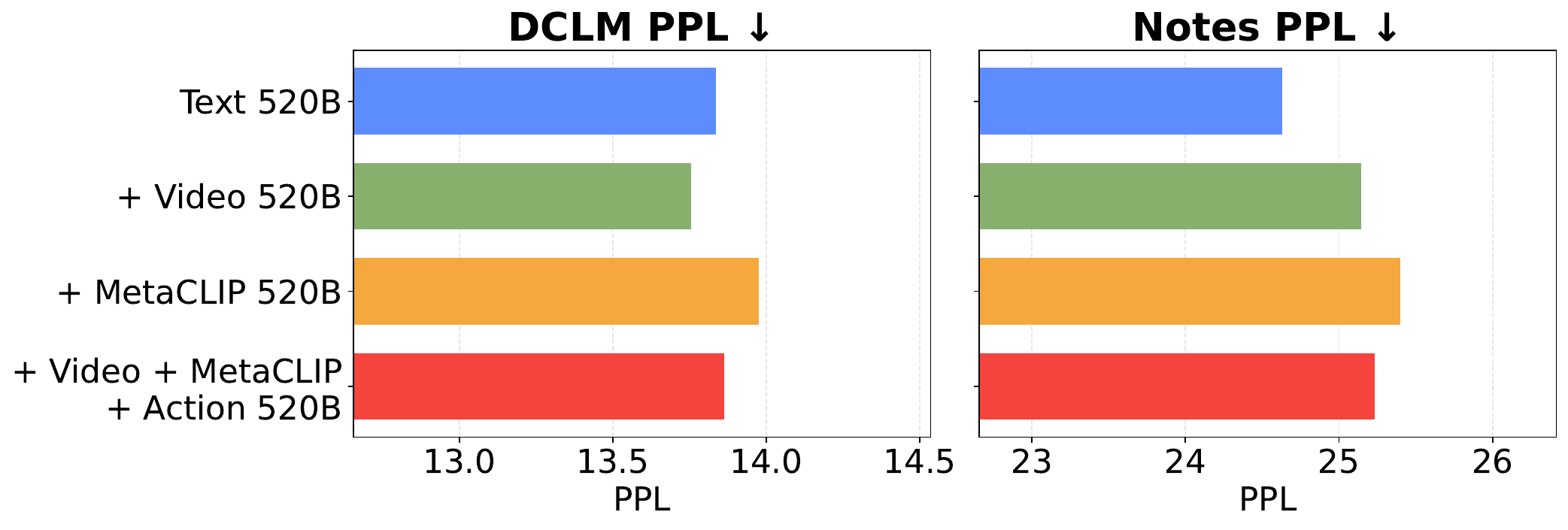} 
        \caption{\textbf{Visual data does not compete with text.} 
        \mix{Text + Video} matches the text-only PPL on DCLM, suggesting that raw visual data is compatible with language modeling. 
        }
        \label{fig:data_ablation_text}
    \end{minipage}
    \hfill 
    \begin{minipage}[b]{0.49
    \textwidth}
        \centering
        \includegraphics[width=\linewidth]{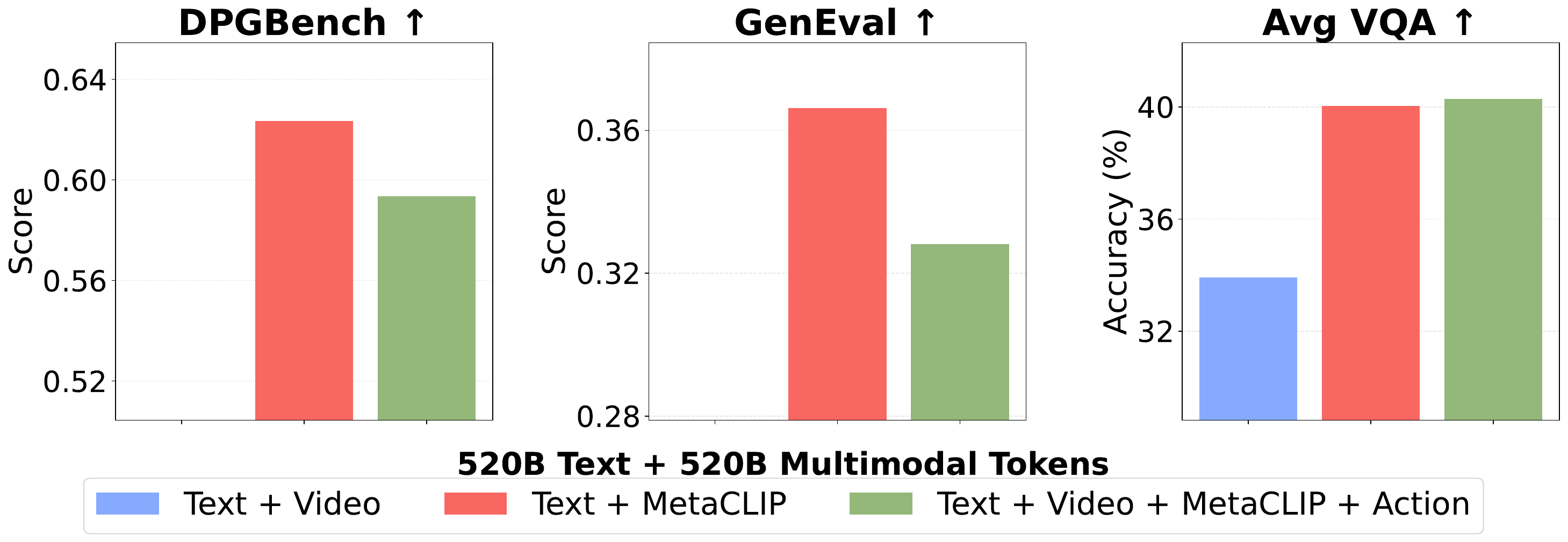}
        \caption{\textbf{I/T data enables visual capabilities.} I/T pairs are essential for visual understanding and generation. \mix{Text + Video} contains no I/T data, so image generation is omitted.
        }
        \label{fig:data_ablation_gen}
    \end{minipage}
\end{figure*}
The premise of unified multimodal pretraining is to leverage all available data. However, it is unclear how each data type contributes to or interferes with the final model. To better understand this, we study three representative data mixtures (see
\Cref{appendix: data sources} for details): \textbf{1)}~\mix{Text + Video} (raw video without text annotations), \textbf{2)}~\mix{Text + MetaCLIP} (I/T pairs) and \textbf{3)}~\mix{Text + Video + MetaCLIP + Action} (all the above + action-conditioned videos). All multimodal models are trained on $\approx$1T tokens (520B text + 520B multimodal) and are compared to a text-only baseline trained on 520B text tokens.

\paragraph{Vision data minimally impacts text performance.} In \Cref{fig:data_ablation_text}, we find that \mix{Text + Video} achieves the best perplexity of all data mixtures on both the DCLM val set and the in-house Notes corpus. On DCLM, \mix{Text + Video} even outperforms the text-only baseline, suggesting that video data is at least compatible with, and possibly beneficial for language modeling. This also suggests that vision itself is not the primary cause of modality competition. On the other hand, \mix{Text + MetaCLIP} achieves the worst perplexity of all data mixtures. \mix{Text + Video + MetaCLIP + Actions} only slightly degrades compared to the text-only baseline, suggesting that video + action trajectories are also complementary to text. We hypothesize that text degradation stems from a shift in the text distribution from introducing image captions, and further investigate this in \Cref{sec:ablation_image_text_distribution}. 

Second, we observe that across all data mixtures, perplexity relative to the text-only baseline degrades on the more out-of-distribution Notes corpus, but the relative trends remain the same. This suggests that multimodal pretraining may introduce a minor trade-off in text generalization. 

\paragraph{I/T data enables visual generation and understanding.} Consistent with prior work~\citep{ tong2024cambrian, InternViT}, in \Cref{fig:data_ablation_gen}, we find that I/T data is essential for acquiring understanding and generation capabilities. Further, our results on VQA demonstrate that understanding tasks benefit significantly from broader data diversity, with the full data mixture surpassing all other combinations of data.

\subsection{Image-Text Distribution}
\label{sec:ablation_image_text_distribution}

In \Cref{sec:ablation_pretrain_data}, we observed that text perplexity slightly degrades when I/T data is introduced (e.g.\ \mix{Text + MetaCLIP}). We hypothesize this stems from the distributional gap between pretraining text and image captions. We study three types of image-text data sources: web-crawled data (MetaCLIP), synthetic captions (MetaCLIP Recaption), and high-aesthetic images (Shutterstock/SSTK).

\begin{figure}[t]
  \centering
  \includegraphics[width=\textwidth]{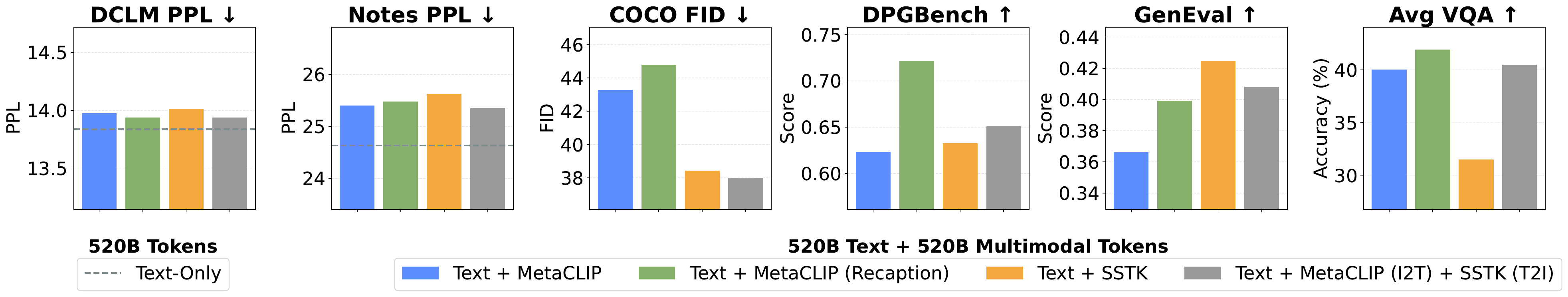}
  \caption{
    \textbf{I/T data sources exhibit complementary trade-offs.} Each source of I/T data has distinct strengths across text perplexity, generation, and VQA. Combining MetaCLIP (I2T) with Shutterstock (T2I) captures the strengths of both, suggesting that strategically using different I/T sources can outperform using a single source of I/T data.}
  \label{fig:text_distribution}
\end{figure}

As quantified in \Cref{tab:text_cosine_distance}, synthetic recaptions and in-house data exhibit a significantly larger cosine distance from the pretraining corpus (DCLM) than standard MetaCLIP captions, directly correlating with the observed degradation in language perplexity, especially on Notes (\Cref{fig:text_distribution}).

\begin{center}
    \small
    \setlength{\tabcolsep}{6pt}
    \begin{tabular}{lccc}
    \toprule
     & MetaCLIP & MetaCLIP (Recap) & SSTK \\
    \midrule
    Cosine Distance ($\downarrow$) & 0.196 & 0.286 & 0.215 \\
    \bottomrule
    \end{tabular}
    \captionof{table}{Cosine distance of image captions to DCLM. See \Cref{appendix:cosine_distance} for details.}
    \label{tab:text_cosine_distance}
\end{center}

Different I/T sources affect visual understanding and generation differently: recaptioned data improves VQA, while high-aesthetic data (SSTK) improves generation quality (\Cref{fig:text_distribution}). Consequently, we decouple the data by objective, using MetaCLIP for image-to-text and SSTK for text-to-image. This captures the strengths of both across all metrics, and suggests that data can be independently chosen to target specific capabilities.

\subsection{Synergy in Multimodal Pretraining}
\label{sec:data_synergy}

\begin{figure}[t]
    \centering
    \begin{minipage}[b]{0.49\textwidth}
        \centering
        \includegraphics[width=\linewidth]{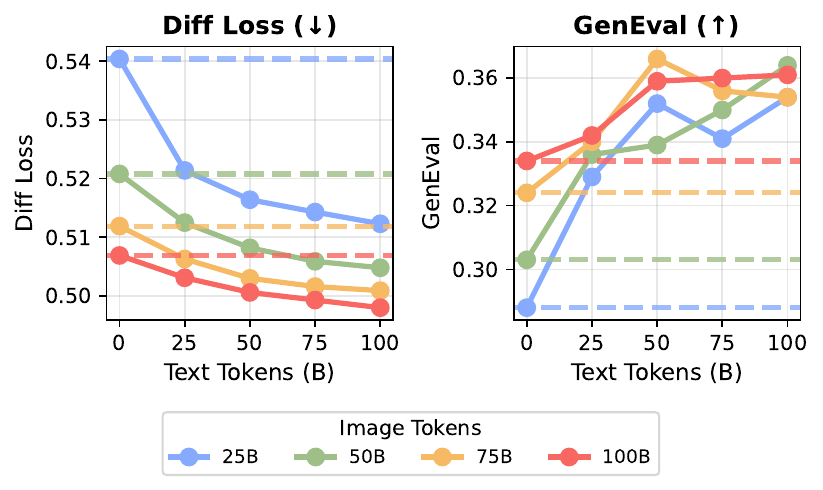}
\caption{\textbf{Multimodal co-training exceeds unimodal performance.} Across all token budgets, adding text tokens to a fixed vision budget consistently improves generation quality (\protect\tikz[baseline=-0.5ex]\protect\draw[thick] (0,0) -- (0.4,0);), surpassing the multimodal-only baseline (\protect\tikz[baseline=-0.5ex]\protect\draw[thick,dashed] (0,0) -- (0.4,0);).}
        \label{fig:data_synergy}
    \end{minipage}
    \hfill 
    \begin{minipage}[b]{0.49\textwidth}
        \centering
        \includegraphics[width=\linewidth]{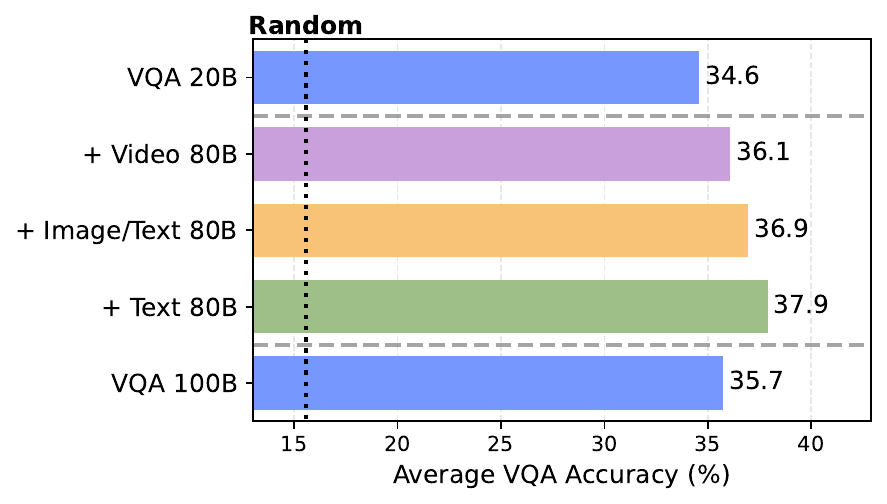}
        \caption{\textbf{General pretraining outperforms specialized scaling.} Supplementing 20B VQA tokens with general-purpose data (text, video, or image-text) yields higher accuracy than training on 100B VQA tokens alone.
}
        \label{fig:vqa_data_ablation}
    \end{minipage}
\end{figure}

We observed in \Cref{sec:ablation_pretrain_data} that adding diverse multimodal data in equal proportion to text minimally impacts text performance. We now explore whether multimodal pretraining can actively \textit{benefit} any capabilities.

\paragraph{Vision and language are complementary.}

We measure how each modality affects the other by training models across all combinations of $\{0, 25, 50, 75, 100\}$B text and multimodal tokens (\Cref{fig:data_synergy}). 
Adding text tokens to a fixed vision budget consistently improves diffusion loss and GenEval score, clearly surpassing the multimodal-only baseline. We also measure text performance (\Cref{appendix: text perf data composition}) and confirm that vision minimally impacts language, while language helps vision. This benefit likely arises because image generation benchmarks (e.g. GenEval) are text-conditioned, so improved language modeling improves text-to-image alignment.

\paragraph{Synergy extends to visual understanding.}

Motivated by this, we ask whether the same synergy extends to visual understanding, or if VQA performance primarily requires domain-specific data. We train models on 20B VQA tokens supplemented with 80B of heterogeneous data (video, MetaCLIP, or text), and compare against both a 20B VQA-only baseline and a 100B VQA-only baseline. As shown in \Cref{fig:vqa_data_ablation}, all mixed variants outperform even the 100B baseline, despite using 5$\times$ less in-domain data. This shows that simply scaling task-specific data is inferior to diverse pretraining, which provides a stronger foundation. The benefit of language data aligns with prior findings~\citep{han2025learning}, and notably we also observe that even out-of-domain data such as unlabeled video improves VQA performance.

We further test this at scale by comparing pretraining on 520B text + 520B multimodal tokens (video, MetaCLIP, action) against a text-only baseline, then finetuning both on VQA following~\citet{tong2024cambrian} (see \Cref{appendix:vqa_details}). As shown in \Cref{fig:native_vs_finetune}, multimodal pretraining improves VQA over text-only pretraining across a variety of visual representations. Notably, semantic encoders such as SigLIP 2 consistently outperform VAE-based encoders, reinforcing our findings in \Cref{sec:ablation_visual_representation}. 

\begin{figure}[h]
  \centering
  \includegraphics[width=\linewidth]{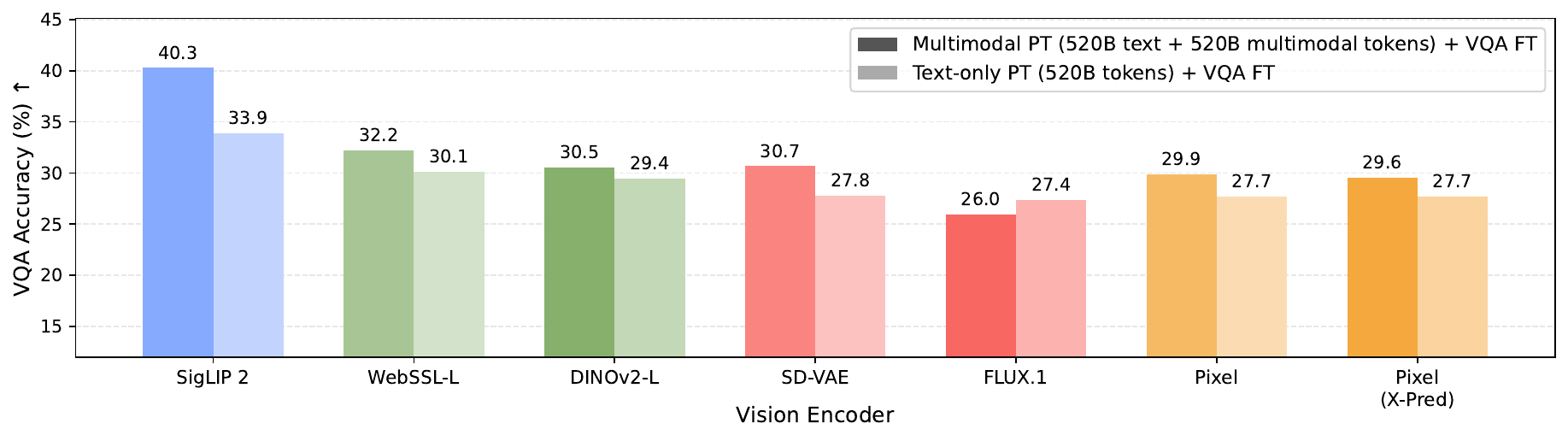}
  \caption{\textbf{Multimodal pretraining is better for VQA.} Multimodal pretraining (dark bars) consistently outperforms text-only baselines (light bars) after finetuning on VQA.}
  \label{fig:native_vs_finetune}
\end{figure}

\suggestion{2}{Train with multimodal data (e.g. video, image-text). Visual data does not degrade language modeling, and diverse pretraining yields synergy for downstream tasks such as world modeling and VQA.}

\section{Towards World Modeling in Unified Multimodal Models}
\label{sec:world_modeling}

\begin{figure}
\centering
\includegraphics[width=0.8\linewidth]{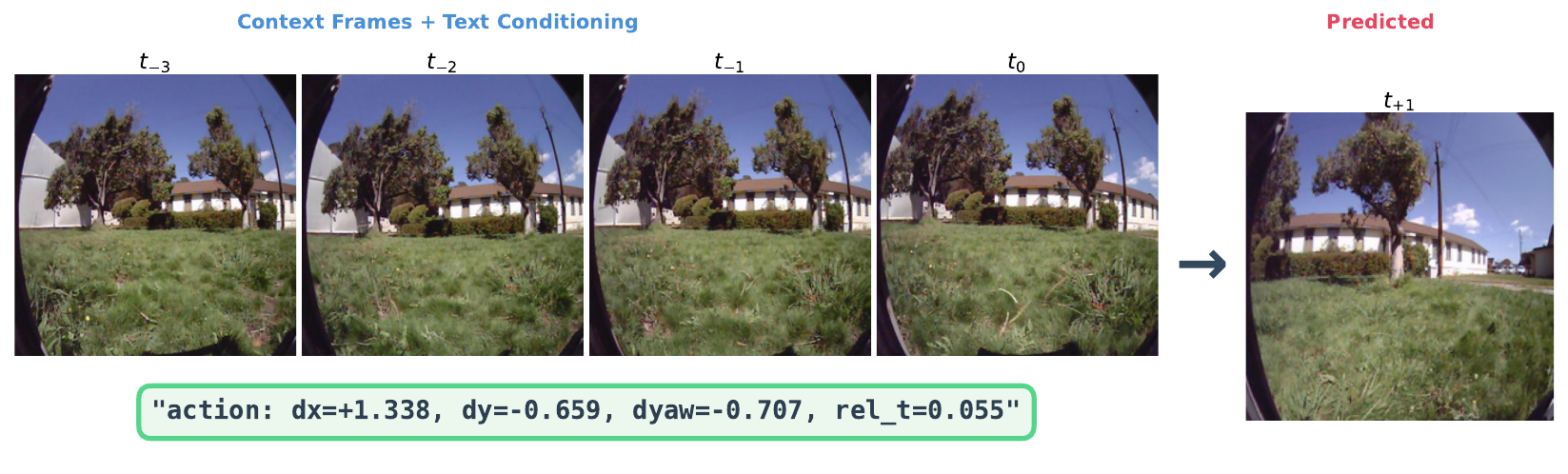}
\caption{ \textbf{Navigation actions are encoded directly as text tokens.} We utilize data from Navigation World Model (NWM), where a training sequence consists of four context frames and a navigation action formatted as natural language text. The model is trained to predict the resulting target view. }\label{fig:nwm_data_example}
\vspace{-0.3em}
\end{figure}
\begin{figure}[t]
\centering
\begin{minipage}[b]{0.49\textwidth}
    \centering
    \includegraphics[width=\linewidth]{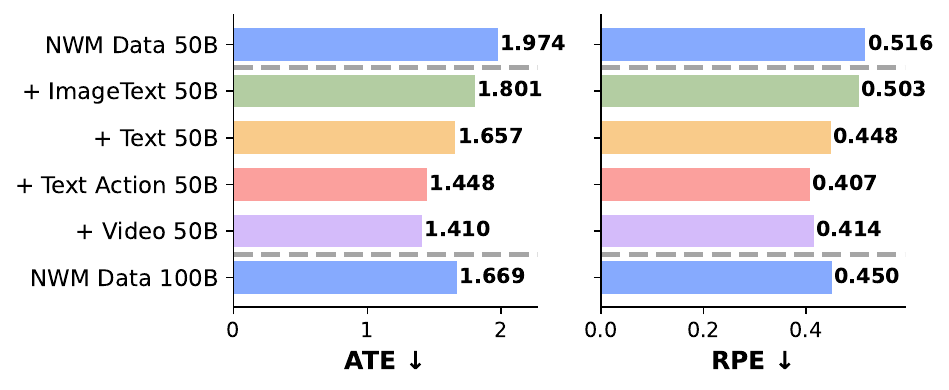}
    \caption{\textbf{Unsupervised video data drives world modeling performance.} Supplementing the NWM baseline with general-purpose video data yields the best performance.
    }\label{fig:nwm_composition}
\end{minipage}
\hfill
\begin{minipage}[b]{0.49\textwidth}
    \centering
    \includegraphics[width=\linewidth]{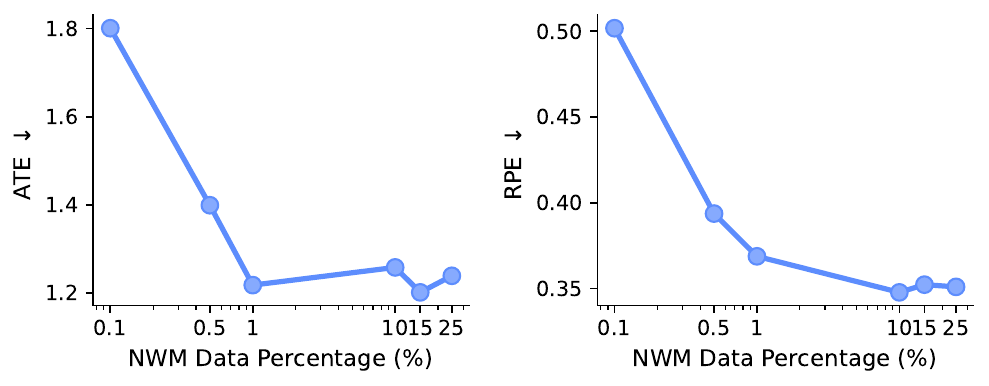}
    \caption{\textbf{World modeling transfers with minimal alignment.} Performance saturates at 1\%, suggesting that the core capability is mostly acquired from general pretraining.
    }\label{fig:nwm_ratio}
\end{minipage}
\end{figure}
Motivated by the observation that language and vision are complementary, and that multimodal pretraining benefits VQA, we then explore whether multimodal models can be extended to do world modeling~\citep{118723, worldmodel, lecun2022path}, without any architectural changes. We adopt the setting of Navigation World Model (NWM)~\citep{bar2025navigation} where the task is to predict the next visual state, conditioned on a given context state and navigation action ($\text{State}_{t} + \text{Action} \rightarrow \text{State}_{t+1}$).

However, unlike NWM which encodes navigation actions (translation and rotation deltas) as specialized continuous vectors, we directly represent actions as standard text tokens (i.e. numerical strings). This allows us to frame the task as \imtok + \texttok$\rightarrow$ \imtok prediction within our unified multimodal model. See \Cref{fig:nwm_data_example} for an example. Unlike NWM, we do not introduce any action-specific adapters nor modify the architecture.

\subsection{NWM Data and Protocol}
We follow the data and evaluation protocols established by NWM~\citep{bar2025navigation}. The model is trained on egocentric robot navigation datasets (e.g. SCAND, RECON)~\citep{shah2021rapid, karnan2022scand}, where each sample consists of 4 context frames, a navigation action, and the resultant target frame. For evaluation, we perform zero-shot planning using the Cross-Entropy Method (CEM)~\citep{rubinstein1997optimization} with $N=120$ samples over an 8-step horizon, optimizing action sequences to minimize the LPIPS distance~\citep{lpips} between the predicted final frame and the goal image, and report the absolute trajectory error (ATE) and relative pose error (RPE). See details in \Cref{appendix:world_model}.

\subsection{World Modeling Emerges from Multimodal Pretraining}
Does effective world modeling stem primarily from domain-specific navigation data, or does it emerge from broader multimodal capabilities? To test this, we compare multimodal models trained on 50B NWM tokens and 50B multimodal data (text, MetaCLIP, videos with text annotations, or video), against a baseline trained on 50B tokens of only NWM data.

The results in \Cref{fig:nwm_composition} mirror the VQA results in \Cref{fig:vqa_data_ablation}. Scaling domain-specific NWM data from 50B to 100B tokens yields moderate improvements in ATE and RPE, but multimodal pretraining provides better results. Specifically, adding pure video yields the biggest gain, but all other modalities including MetaCLIP and text also help. This suggests that world modeling relies more on capabilities acquired from multimodal pretraining, rather than domain-specific data. This mirrors findings from earlier work~\citep{dreamerv4}.

\paragraph{World modeling transfers from general training.}
To further disentangle the origin of world modeling capabilities, we conduct an ablation study varying the ratio of NWM data while keeping the total training budget fixed at 200B tokens (see \Cref{appendix:world_model} for details). As shown in \Cref{fig:nwm_ratio}, performance saturates rapidly with respect to domain data volume. We observe that the model reaches competitive performance with as little as 1\% in-domain data, with negligible gains observed at higher ratios.

Collectively, our findings reinforce our hypothesis that abilities such as navigation and VQA are primarily learned from general pretraining and only require minimal in-domain data.

\begin{figure}[t]
\centering
\includegraphics[width=0.8\linewidth]{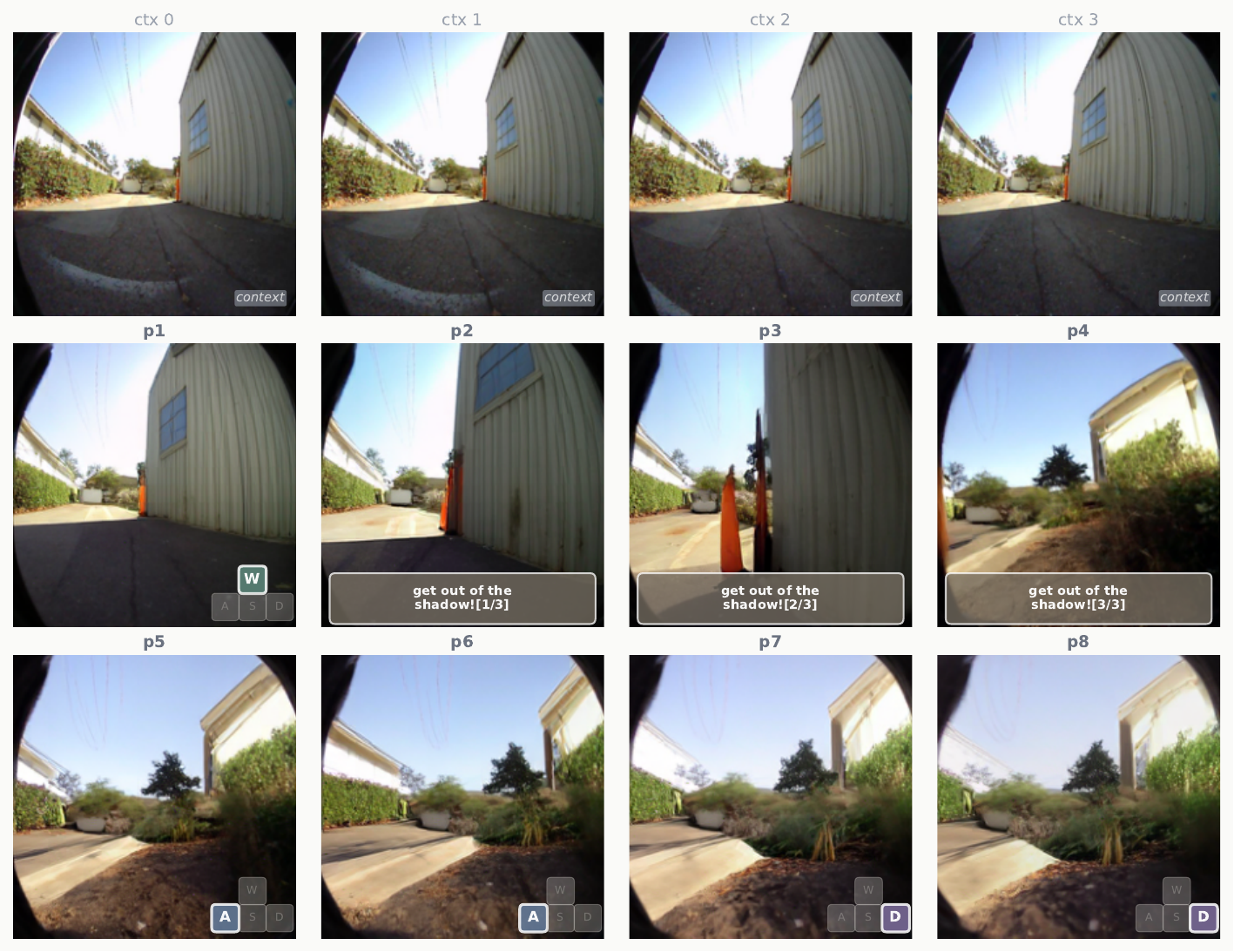}
\caption{
    \textbf{Zero-shot model controllability via natural language.}
    Given four context frames (\texttt{ctx0}--\texttt{ctx3}), the model predicts eight future frames (\texttt{p1}-\texttt{p8}) conditioned on WASD-style or natural language navigation actions.
  }\label{fig:nwm_qualitative}
\end{figure}

\subsection{Qualitative Rollouts with Free-form Language as Actions}
We visualize the model's rollout capabilities by providing four context frames and predicting subsequent frames autoregressively. We first predefine four actions corresponding to standard WASD controls (Forward {\small\fbox{\texttt{W}}}, Left {\small\fbox{\texttt{A}}}, Backward {\small\fbox{\texttt{S}}}, Right {\small\fbox{\texttt{D}}}); we refer readers to \Cref{appendix:world_model} for the specific numerical values of these actions. As a consequence of our multimodal pretraining, we can also condition on free-form natural language prompts as actions (e.g. ``get out of the shadow!''), which are out-of-distribution. See \Cref{appendix:more_navigation_examples} for more qualitative examples.

\Cref{fig:nwm_qualitative} displays a generated sequence produced by controlling the model with these inputs. We observe that the generated frames maintain visual consistency while faithfully executing the commanded actions. The model is also able to generate trajectories directly based on natural language commands. This language-driven navigation emerges as a \textit{zero-shot} capability, demonstrating how semantic understanding successfully transfers from the broader multimodal pretraining.

\suggestion{3}{Unified multimodal pretraining unlocks world modeling. Represent actions as text without architectural changes; capabilities emerge via general training with minimal domain-specific data.}

\section{Unified Multimodal Architecture Design}
\label{sec:arch_design}

We show in \Cref{fig:ffn_shared_vs_specific} that simply replacing shared FFNs with modality-specific FFNs is highly effective, demonstrating the potential of modest capacity separation. However, modality-specific FFNs separate capacity between two modalities evenly, which may not reflect the ideal capacity allocation.

We now explore whether Mixture-of-Experts (MoE) can learn this separation dynamically by decoupling total capacity from active compute. We study the MoE design space (\Cref{sec:moe}), the specialization patterns that emerge (\Cref{subsec:moe_analysis}), and alternative design choices (\Cref{sec:arch_ablations}).

\subsection{Exploring MoEs for Unified Multimodal Models}
\label{sec:moe}
Mixture-of-Experts (MoE) generalizes fixed modality-specific FFNs by learning a dynamic decomposition of the FFN layers, routing each token to a subset of specialized experts~\citep{shazeer2017outrageously, fedus2022switch}.
Recent work~\citep{deepseekv2, deepseekv3, gptoss, kimik2, qwen3} has demonstrated the scalability of MoE for language modeling. Here, we study its design space for unified multimodal pretraining.
Unless otherwise noted, all MoE experiments are trained for 57B tokens on a 50/50 mixture of DCLM text and Shutterstock image data.

\paragraph{Granularity.}
\label{subsec:moe_granularity}

We first analyze the role of expert \emph{granularity} $G$, which we define relative to model dimension $d_\text{model}$ and expert dimension $d_\text{expert}$ as
\begin{equation}
    G = \frac{4d_\text{model}}{d_\text{expert}}.
\end{equation}
We sweep $G \in \{1, 4, 16, 32, 64\}$ at fixed active compute: as $G$ increases, the number of active experts increases while the dimension of each expert shrinks proportionally. $G{=}1$ corresponds to a baseline of 16 large experts ($d{=}8192$) with Top-1 routing, while $G{=}64$ yields 1024 small experts ($d{=}128$)\footnote{Due to implementation constraints in TorchTitan~\citep{torchtitan}, we use 1008 experts for the $G=64$ setting.} with Top-64 routing.

\begin{figure}[t]
  \centering
  \includegraphics[width=\linewidth]{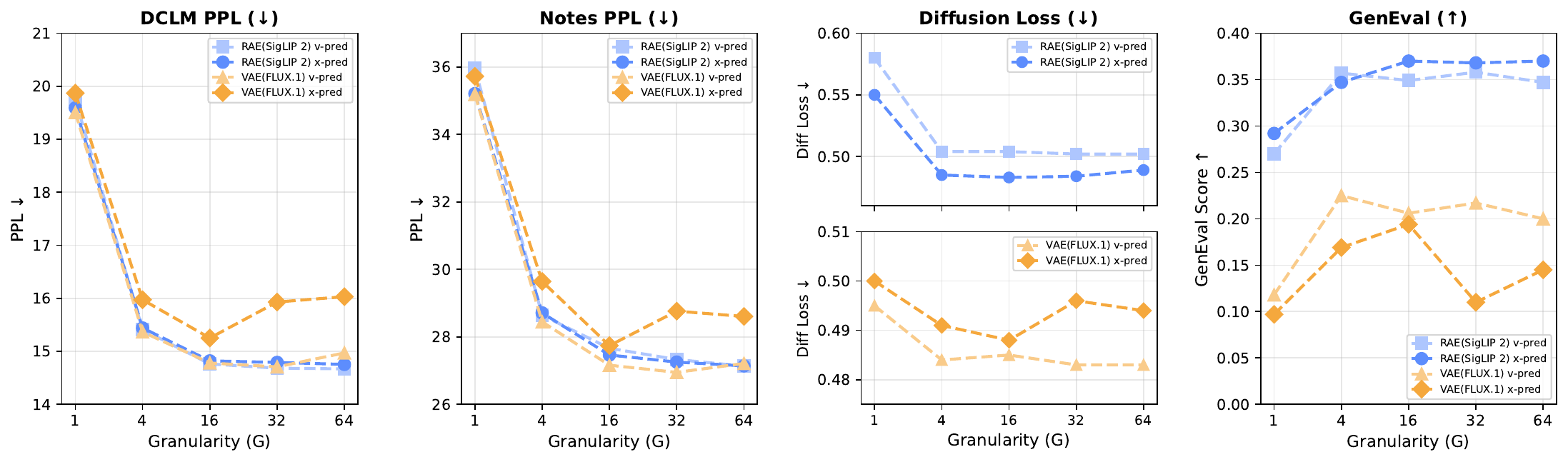}
  \caption{
    \textbf{Fine-grained experts improve performance; $x$-pred benefits high-dimensional diffusion.} 
    Increasing granularity improves all metrics up to $G{=}16$.
    For RAE (SigLIP 2), $x$-pred (solid) outperforms $v$-pred (dashed) on image generation; for VAE (FLUX.1), $v$-pred is better.
  }\label{fig:moe_granularity}
  \vspace{-0.5em}
\end{figure}

We test two visual representations---RAE (SigLIP 2) and VAE (FLUX.1)---to understand how the visual latent space affects MoE efficacy. We also test two diffusion prediction targets: $x$-pred~\citep{li2025jit} and $v$-pred~\citep{fm}. Because finer-grained experts have a smaller dimension which may fall below the rank of the visual latent space, we hypothesize that $x$-pred can mitigate this by leveraging the manifold assumption. All models have 13.5B total parameters (1.5B active); see \Cref{appendix:moe_models} for detailed configurations.

As shown in \Cref{fig:moe_granularity}, \textbf{higher granularity is critical}: increasing $G$ from 1 to 16 substantially improves both language and vision modeling, as many smaller experts enable better capacity allocation across modalities. Notably, the two modalities saturate at different granularities---vision at $G=4$, language at $G=16$---suggesting that language benefits more from fine-grained routing. Gains plateau beyond $G=32$, so we adopt $G=16$ for subsequent experiments.

\paragraph{Prediction target depends on visual representation.}
The optimal diffusion prediction target varies with the visual representation (\Cref{fig:moe_granularity}, solid vs.\ dashed). For RAE (SigLIP 2), $x$-pred~\citep{li2025jit} consistently outperforms $v$-pred on image generation across all granularities, with minimal impact on language. For VAE (FLUX.1), however, $x$-pred causes text perplexity to spike as granularity increases, while $v$-pred scales cleanly. This suggests that $x$-pred is less stable for low-dimensional representations as the expert dimension shrinks, while high-dimensional representations can better leverage higher expert granularity.

\paragraph{Sparsity.}
\label{subsec:moe_sparsity}

\begin{figure}[t]
  \centering
  \includegraphics[width=\linewidth]{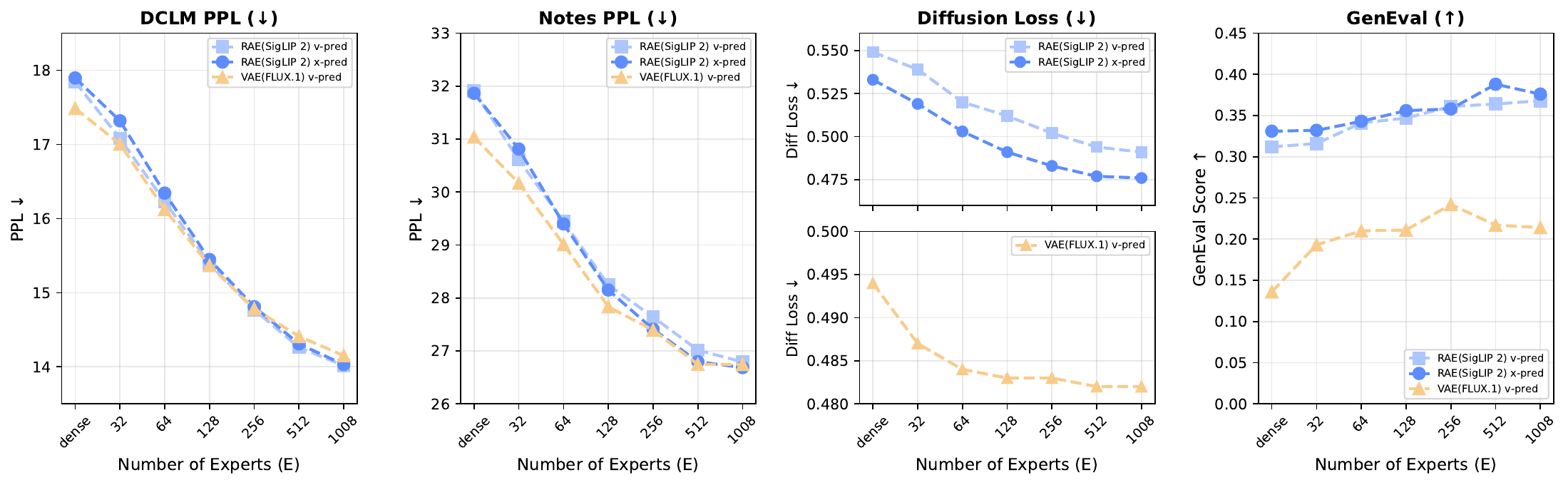}
  \caption{
    \textbf{Sparsity scales multimodal performance.} 
    At fixed \emph{active} compute (16 experts), increasing \emph{total} expert count from 32 to 1008 (and thus total parameter count) consistently improves both language (lower PPL) and vision (lower diffusion loss, higher GenEval score).
  }\label{fig:moe_sparsity}
  \vspace{-0.5em}
\end{figure}

MoE decouples total capacity from active compute; the number of active experts is fixed while the total expert pool grows, thus efficiently increasing model capacity without additional training or inference cost. Using the optimal $G=16$ from the granularity sweep, we fix the active budget (16 experts, hidden dim 512) and increase the total expert count from 32 to 1008, reducing the active ratio from 50\% to 1.6\%. We also compare against a 1.5B dense baseline with the same compute budget.

As shown in \Cref{fig:moe_sparsity}, both modalities benefit consistently from increased sparsity: text PPL decreases and visual generation improves as total experts grow from 32 to over 1000---all at fixed compute. This confirms that MoE, which is already widely used for language, can also extend to unified multimodal pretraining.

\begin{figure}[t]
  \centering
  \includegraphics[width=\linewidth]{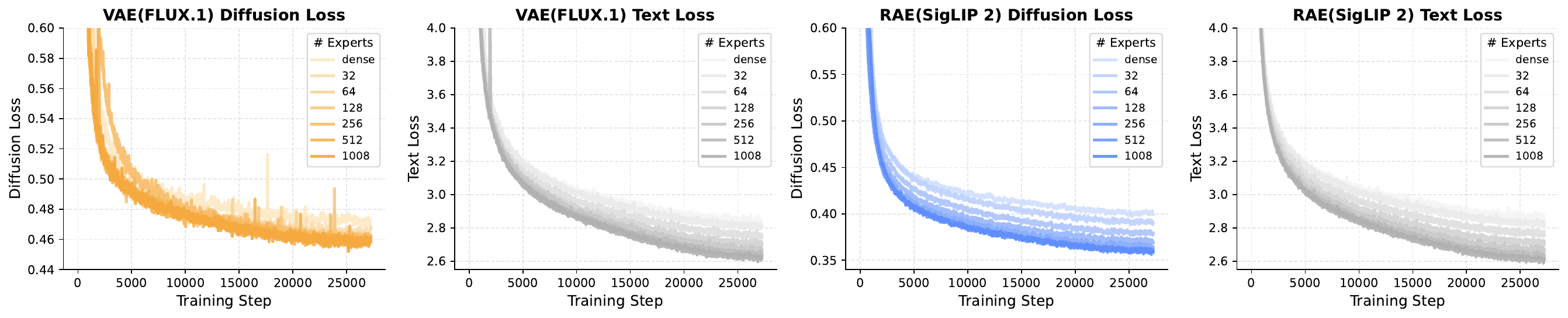}
  \caption{\textbf{Semantic representations scale better with sparsity.} We fix 16 active experts and increase the total expert count. Text loss improves consistently regardless of the vision encoder, but diffusion loss saturates for VAE (FLUX.1) while continuing to improve for RAE (SigLIP 2).}
  \label{fig:moe_loss}
\end{figure}

The training loss (\Cref{fig:moe_loss}) reveals a further distinction between visual representations: for RAE (SigLIP 2), both text and diffusion loss continue to improve with more experts, while for VAE (FLUX.1), diffusion loss saturates, suggesting that semantic representations benefit more from sparsity.

\paragraph{Expert design choices.}
Given that language and vision respond differently to granularity, we explore whether modality-specific expert designs help. Previous work~\citep{dai2024deepseekmoe,deepseekv2} in language modeling shows that designating certain experts as always active (``shared experts'') improves performance. We compare three configurations, all with 256 total experts and 16 active:
\begin{enumerate}
    \item \textbf{No Shared Expert:} Standard routing where the top-16 experts are selected from the full pool of 256.
    \item \textbf{Global Shared Expert:} 1 expert is designated as shared (always active), and the router selects the top-15 from the remaining 255.
    \item \textbf{Per-Modality Shared Expert:} We designate 1 shared expert for text and 1 for vision. For any given token, only the corresponding modality-specific shared expert is activated, along with the top-15 routed from the remaining 254.
\end{enumerate}

\begin{table}[h]
\centering
\caption{\textbf{Per-Modality Shared Experts outperform Global Shared Experts.} We compare standard Top-16 routing against Global and Per-Modality shared expert strategies. Per-modality yields the best results.}
\label{tab:moe_shared_expert}
\begin{tabular}{lcccc}
\toprule
Configuration & DCLM PPL $\downarrow$ & Notes PPL $\downarrow$ & Diff.\ Loss $\downarrow$ & GenEval $\uparrow$ \\
\midrule
No Shared Expert       & 14.802 & 27.392 & 0.484 & 0.360 \\
Global Shared Expert   & 14.794 & 27.249 & 0.483 & 0.364 \\
Per-Modality Shared Expert & \textbf{14.785} & \textbf{27.161} & \textbf{0.483} & \textbf{0.367} \\
\bottomrule
\end{tabular}
\end{table}

As shown in \Cref{tab:moe_shared_expert}, while a global shared expert improves over the baseline, per-modality shared experts yield the best performance across all metrics. This suggests that different modalities have distinct capacity needs that benefit from dedicated computation---a theme we examine in depth in \Cref{subsec:moe_analysis}.

\subsection{Emergent Expert Specialization}
\label{subsec:moe_analysis}
In \Cref{sec:moe}, we explored the MoE design space across granularity, sparsity, prediction targets, and shared experts, establishing that MoE is effective for multimodal pretraining. We now ask: \emph{what does the model learn?}
Unlike fixed modality-specific FFNs (\Cref{sec:method}), MoE imposes no human priors on how capacity is allocated between modalities.
We analyze a 13.5B-parameter MoE model (1.5B active, $G{=}16$, 256 experts, SigLIP 2, $x$-pred) trained on 1T mixed multimodal tokens, examining expert allocation along three dimensions: modality, diffusion timesteps, and visual tasks (see \Cref{appendix:moe_analysis} for methodology).

\begin{figure}[t]
    \centering
    \begin{minipage}[t]{0.49\textwidth}
        \centering
        \includegraphics[width=\linewidth]{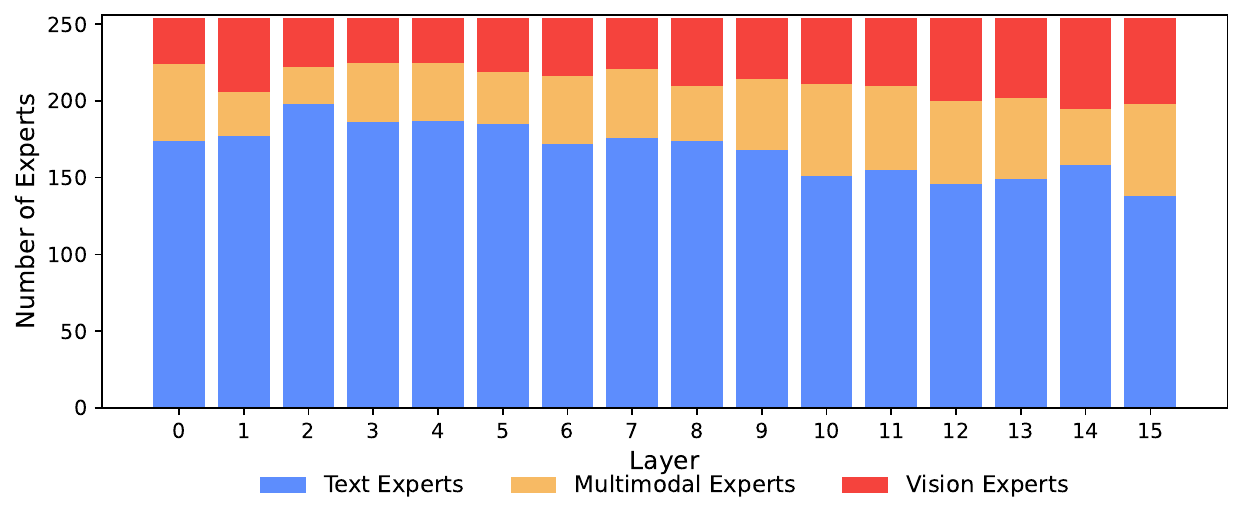}
        \caption{\textbf{Emergent Modality Specialization.} Expert specialization forms naturally. Most experts are text-focused, but later layers contain more vision and multimodal experts.}
        \label{fig:moe_expert_analysis}
    \end{minipage}
    \hfill 
    \begin{minipage}[t]{0.48\textwidth}
        \centering
        \includegraphics[width=0.85\linewidth]{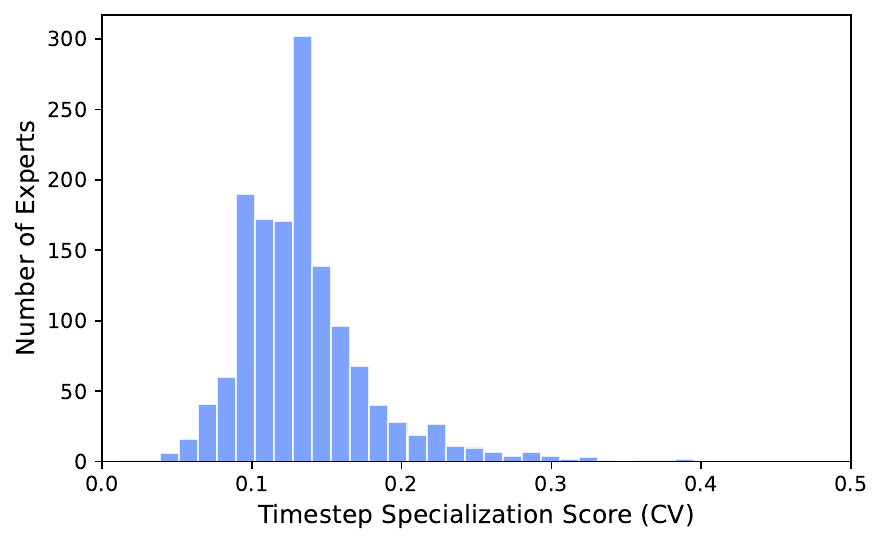}
        \caption{\textbf{Experts do not specialize in diffusion timesteps.} CV scores are distributed near 0.15, indicating vision experts are time-invariant.}
        \label{fig:moe_timestep}
    \end{minipage}
\end{figure}

\paragraph{Modality specialization emerges naturally.}
We classify each expert as Text, Vision, or Multimodal based on its routing preference across held-out DCLM (text) and CC12M (image) tokens.
As shown in \Cref{fig:moe_expert_analysis}, the model allocates significantly more experts to text than to vision, despite the use of an auxiliary load-balancing loss that encourages uniform utilization.
This emergent asymmetry is consistent with the scaling laws we later derive in \Cref{sec:scaling} and mirrors recent findings~\citep{wang2026ernie50technicalreport}: language is parameter-hungry while vision is data-hungry, so the model naturally devotes more dedicated capacity to language via MoE.

We also observe that expert composition evolves across depth: early layers are dominated by text-specific experts, while later layers contain progressively more vision and multimodal experts (\Cref{fig:moe_expert_analysis}).
This suggests that the model learns a separate-then-integrate processing strategy of first processing each modality with dedicated parameters and later fusing them in deeper layers.

\begin{figure}[t]
    \centering
    \includegraphics[width=\linewidth]{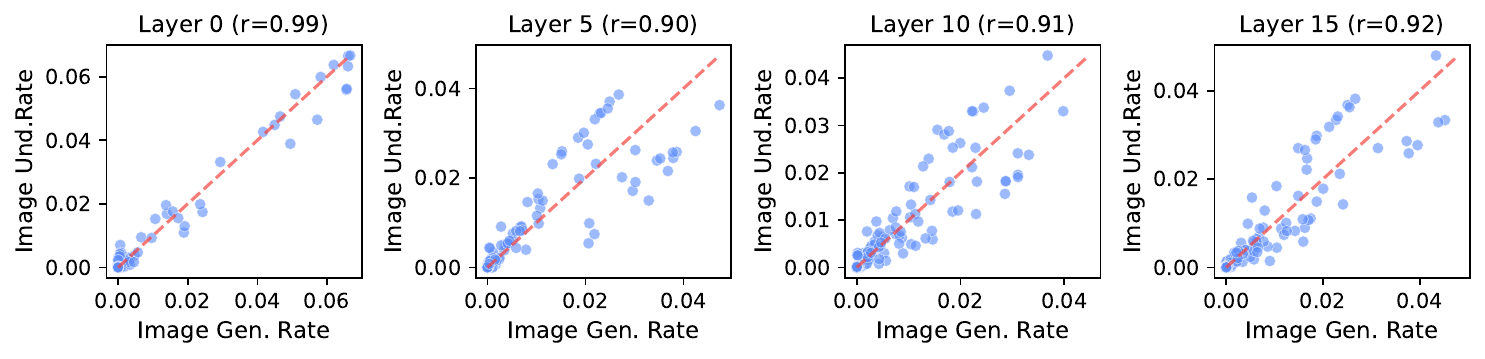}
    \caption{\textbf{Visual generation and understanding share the same experts.} We compare the expert selection rates for image generation (x-axis) versus image understanding (y-axis) across different network depths. We observe very high Pearson correlation ($\geq r=0.90$) across all layers.
    }\label{fig:moe_gen_und}
\end{figure}

\paragraph{Vision experts are general-purpose.}
A natural hypothesis is that vision experts might specialize across the diffusion noise schedule, separating high-noise denoising from fine-grained refinement, as explicitly enforced in architectures like Wan~\citep{wan2025wan}.
We test this by measuring the Coefficient of Variation (CV) of expert selection rates across 10 timestep bins (see \Cref{appendix:moe_analysis}).
As shown in \Cref{fig:moe_timestep}, CV scores cluster near 0.15, indicating minimal timestep specialization---the same experts process the same visual tokens throughout the entire generation trajectory.

Notably, visual understanding and generation share a similar set of experts.
Comparing routing patterns for image-to-text (understanding) versus text-to-image (generation), we observe very high correlation ($r\geq0.90$) across all layers (\Cref{fig:moe_gen_und}): the experts activated for captioning are the same experts activated for denoising.
This suggests that with learnable capacity separation, the model is able to converge to a truly unified visual representation for both tasks---consistent with prior work showing that understanding and generation are mutually beneficial~\citep{tong2024metamorph, li2025manzanosimplescalableunified, uniflow}.

\subsection{Stacking Design Choices}
\label{sec:arch_ablations}

\begin{wrapfigure}{r}{0.445\textwidth}
  \centering
   \vspace{-15pt}
   \includegraphics[width=\linewidth]{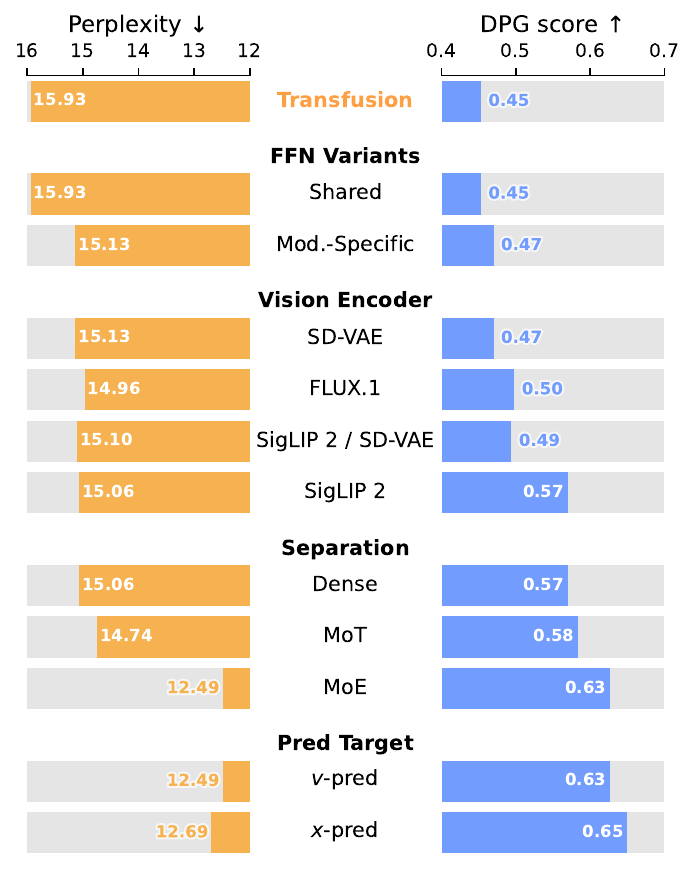}
    \caption{
      \textbf{Modality-specific FFN, SigLIP 2, MoE, and $x$-pred achieve best unified performance.} 
      Starting from a Transfusion baseline, we progressively stack the best design choices for FFN, vision encoder, capacity separation, and prediction target. 
      Each row shows \textcolor{DiTyellow5}{\textbf{Perplexity}~($\downarrow$)} and \textcolor{DiTblue5}{\textbf{DPG~score}~($\uparrow$)}. 
  }\label{fig:progression}
  \vspace{-3em}
\end{wrapfigure}

The preceding sections studied each design axis in isolation: visual representation (\Cref{sec:ablation_visual_representation}), data (\Cref{sec:data_composition}), and MoE configuration (\Cref{sec:moe}). We now stack these choices progressively to verify these changes are complementary, building from a Transfusion baseline toward our final configuration. We also compare against alternative approaches such as Mixture of Transformers (MoT)~\citep{liang2024mixture}. All models are trained with identical hyperparameters, data, token counts, and FLOPs. \Cref{fig:progression} summarizes the results.

\paragraph{FFN structure.}
Replacing the shared FFN with modality-specific FFNs (separate weights for text and vision), as in
LMFusion~\citep{lmfusion}, improves perplexity from 15.93 to 15.13 and DPG score from 0.45 to 0.47, consistent with \Cref{fig:ffn_shared_vs_specific}. Even
this simplest form of capacity separation reduces modality competition at no additional inference cost, since only one FFN is active per token (see \Cref{sec:ablation_ffn_shared_vs_specific}). We adopt modality-specific FFNs as the default for subsequent comparisons, until we explore MoE.

\paragraph{Vision encoder.}
Using modality-specific FFN, we compare four vision encoders: three single-encoder configurations (SD-VAE, FLUX.1, and SigLIP 2 for RAE-style diffusion) and a dual-encoder setup that uses SigLIP 2 for understanding and SD-VAE for generation, as in Bagel~\citep{deng2025bagel} and Janus~\citep{ma2025janusflow, chen2025januspro}. SigLIP 2 significantly outperforms all alternatives, improving DPG score from 0.47 to 0.57 while also reducing perplexity from 15.13 to 15.06. Notably, the dual-encoder baseline falls short of the single SigLIP 2 encoder on both metrics, suggesting that a unified semantic representation is more effective than separate encoding pathways. These results are consistent with \Cref{sec:ablation_visual_representation} and \Cref{sec:scaling_moe}.

\paragraph{Parameter separation strategy.}
Using SigLIP 2 and modality-specific FFN, we compare three parameter separation approaches: dense (no further separation), deep separation via MoT~\citep{liang2024mixture, deng2025bagel}, and data-driven separation via MoE. MoE achieves the best trade-off, reaching the lowest perplexity (12.49) and highest DPG score (0.63), outperforming both the dense model and MoT.

Beyond raw performance, MoE has a structural advantage: capacity allocation between modalities is difficult to determine a priori. Fixed separation cannot adapt to different inputs, whereas MoE learns to route each token independently during training, adapting capacity allocation to the input (\Cref{sec:moe}). This echoes the ``bitter lesson''~\citep{sutton2019bitter}: learning from data generally outperforms hand-crafted designs.
Our conclusions stem from a \textit{from-scratch} pretraining setting. Approaches such as MoT have proven effective when finetuning pretrained dense LLMs~\citep{liang2024mixture, deng2025bagel}. However, as language models increasingly adopt native MoE architectures~\citep{deepseekv3, kimik2}, extending their routing networks with vision and multimodal experts offers a natural path to multimodal finetuning.

\paragraph{Prediction target.}
Finally, we compare $v$-prediction and $x$-prediction objectives within the best architecture (MoE + SigLIP 2). Switching from $v$-pred to $x$-pred further improves DPG score from 0.63 to 0.65, consistent with the analysis in \Cref{sec:ablation_visual_representation,sec:moe}.

\paragraph{Knowledge-informed generation.}
While standard metrics like FID evaluate distributional visual quality, a key promise of unified models is their ability to transfer world knowledge from language into visual generation. To test how our architectural design choices affect this capability, we evaluate the models on the WISE benchmark~\citep{wise}, which probes factual understanding across categories such as science and culture.

\begin{figure}[t]
  \centering
  \includegraphics[width=\linewidth]{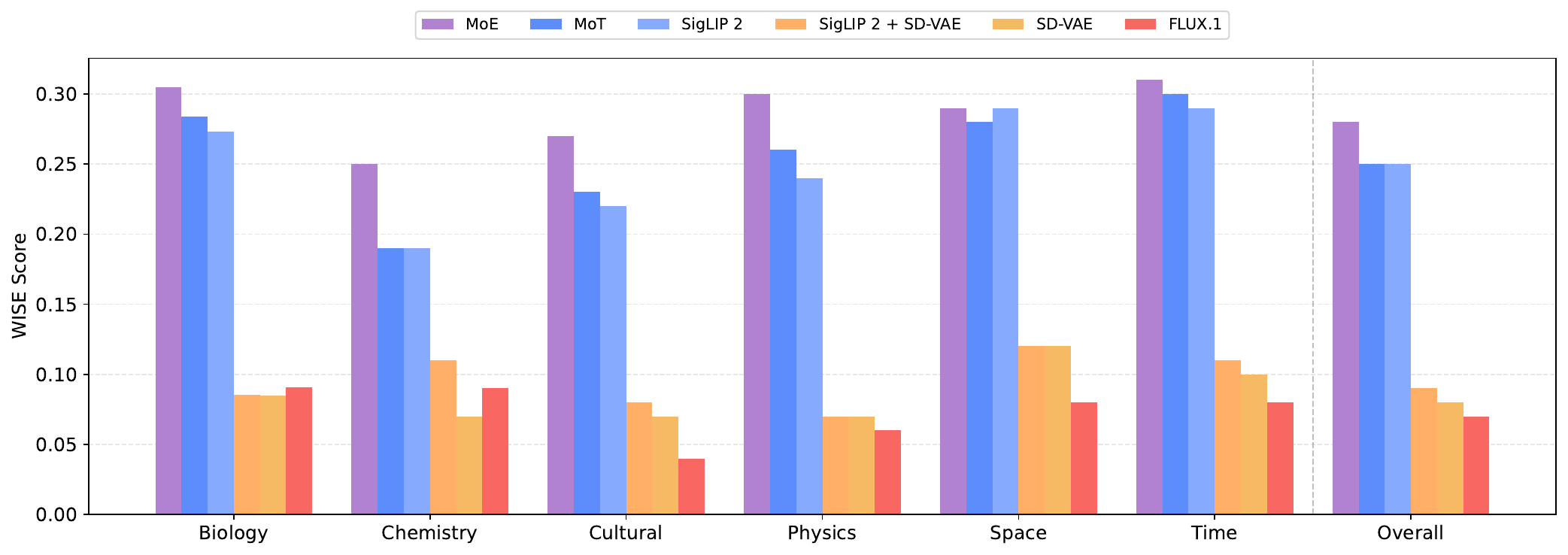}
  \caption{
    \textbf{Semantic encoders and MoE capture more world knowledge.}
    WISE scores across six knowledge categories and overall. Models with semantic encoders (MoE, MoT, SigLIP 2) consistently outperform VAE-based models (SD-VAE, FLUX.1) by 3--4$\times$, with MoE achieving the highest overall score.
  }\label{fig:wise}
\end{figure}

\Cref{fig:wise} reveals two findings. First, the choice of vision encoder is critical for semantic generation: semantic encoders (e.g.,\ SigLIP 2) outperform VAE-based models (SD-VAE, FLUX.1, and dual-encoder) by 3--4$\times$ across all knowledge categories, regardless of the separation strategy. Second, for a fixed encoder, data-driven sparsity is greatly beneficial: MoE outperforms both MoT and the dense baseline, suggesting that learned routing better transfers semantic priors for generation.

\suggestion{4}{Use MoE in unified models: it outperforms hand-crafted separation strategies and naturally learns modality-specific specialization from the data.}

\section{Scaling Laws for Unified Multimodal Models}
\label{sec:scaling}

\begin{figure}[t]
  \centering
    \includegraphics[width=0.98\linewidth]{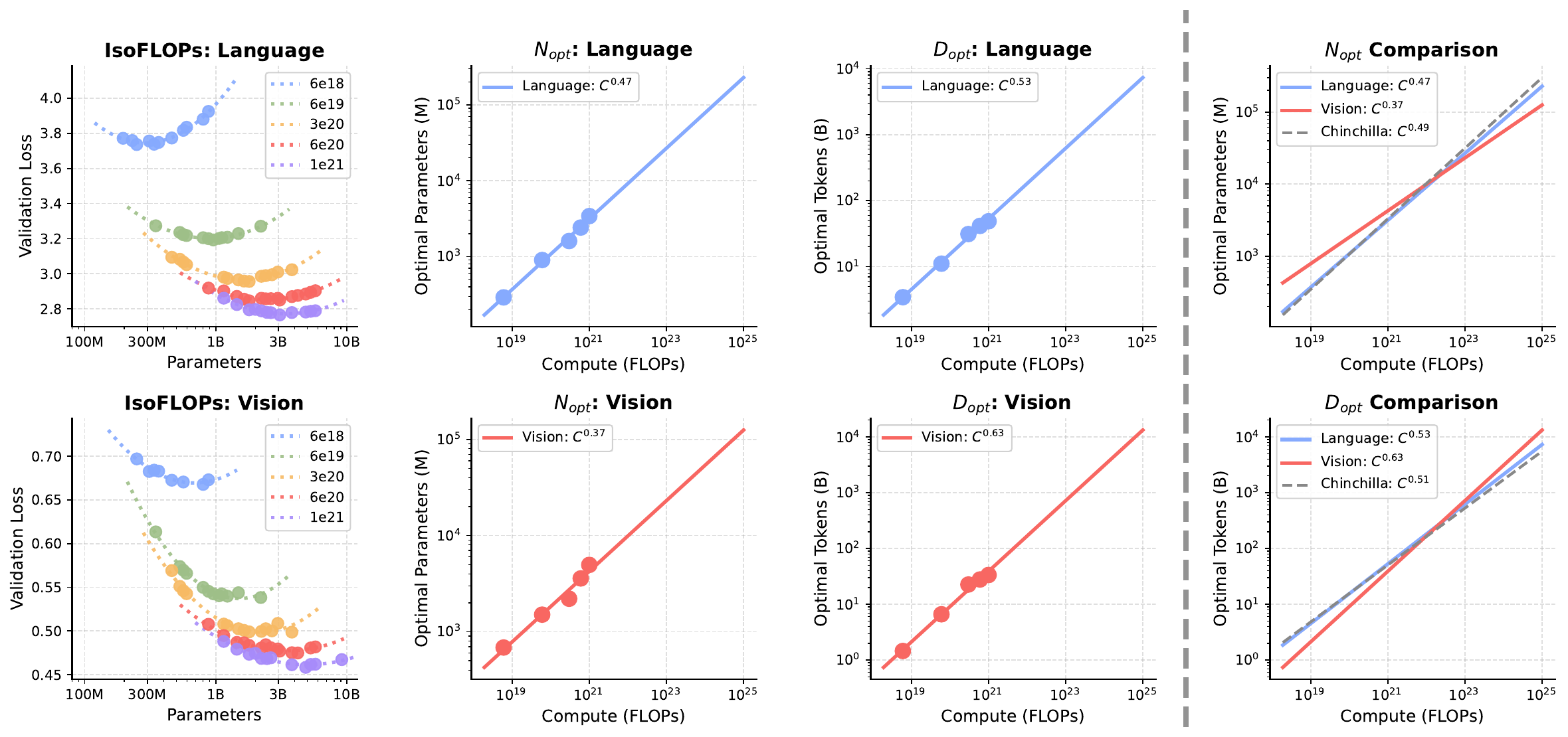}
  \caption{\textbf{Scaling laws for unified dense models.} We compute IsoFLOP curves across compute budgets. We project the optimal model size and token count for each FLOP budget, and find that the vision modality is more data-hungry, while text follows Chinchilla-like scaling.}
  \label{fig:scaling_dense_combined}
\end{figure}

\begin{figure*}[t]
  \centering
  \includegraphics[width=0.98\linewidth]{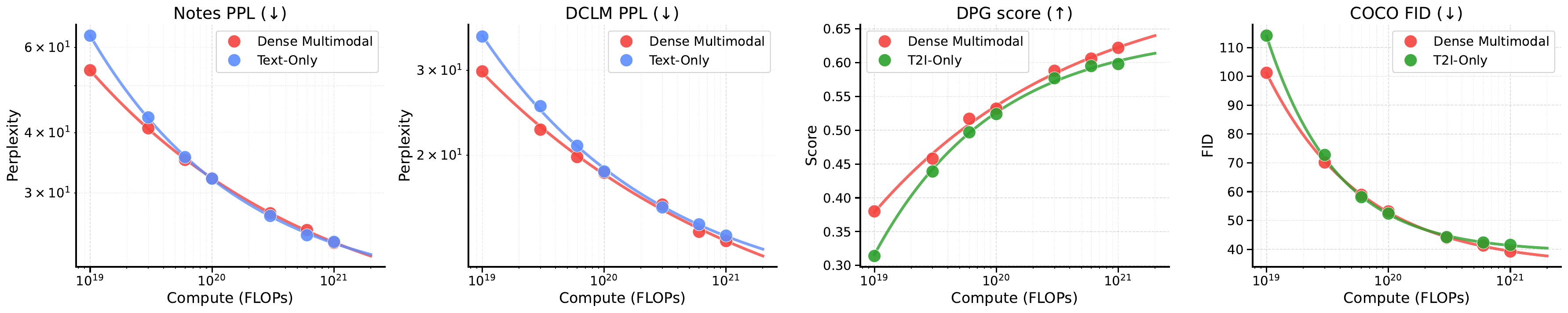}
  \caption{\textbf{Scaling laws and efficiency benchmarks for unified dense models.} We fit $L(C) = A \cdot C^{-\alpha} + E$ to each model class. Dense Multimodal outperforms both the unimodal Text-Only and T2I-Only baselines across all four metrics.}
  \label{fig:scaling_dense}
\end{figure*}

Prior scaling studies primarily examine unimodal language models~\citep{kaplan2020scaling,hoffmann2022chinchilla, tian2025towards}. \citet{shukor2025scaling} studies scaling for multimodal models that only output text. We derive scaling laws for both vision and language, and investigate how architecture influences these trends.

\paragraph{IsoFLOP methodology.}
Following Chinchilla-style analyses~\citep{hoffmann2022chinchilla}, we sweep model sizes $N$ and token counts $D$ at fixed compute budgets $C$. We estimate FLOPs via \texttt{torchtitan}~\citep{torchtitan}, applying $6ND \approx C$ for dense models and using active parameters ($N_{\text{active}}$) for MoE, since each token routes to a subset of experts. We measure validation loss on held-out data (DCLM for language, CC12M for vision) and fit a parabola in log-parameter space to estimate $N_{\text{opt}}(C)$. We then fit a linear regression between $\log N_{\text{opt}}$ and $\log C$, assuming the power-law relationship $N_{\text{opt}} \propto C^a$ and $D_{\text{opt}} \propto C^b$ where $a+b=1$.

\paragraph{Compute-optimal for dense models.}
\label{sec:scaling_dense}

\Cref{fig:scaling_dense_combined} presents dense IsoFLOP results. Fitting power laws $N_{\text{opt}}\propto C^{a}$ and $D_{\text{opt}}\propto C^{b}$ reveals divergent scaling between modalities. For \Language, $a\approx0.47$ and $b\approx0.53$, which is nearly balanced and consistent with Chinchilla scaling laws. For \Vision, $a\approx0.37$ and $b\approx0.63$, indicating that vision is more data-hungry. These distinct exponents show that no single compute-optimal trend governs unified models. This asymmetry creates a practical dilemma at
scale: the data demands of both modalities cannot be simultaneously satisfied. Relating optimal tokens directly to model size ($D_{\text{opt}} \propto N_{\text{opt}}^{b/a}$), the ratio of required vision data to language data grows as $\mathcal{O}(N^{0.57})$. From a 1B parameter baseline, this relative demand increases by $14\times$ at 100B parameters and $51\times$ at 1T. Although recent language models~\citep{grattafiori2024llama3, deepseekv3} are often heavily overtrained, this widening gap forces a compromise: either under-train the vision modality or expend excessive compute over-training language. At the 1T parameter scale, satisfying vision's optimal data requirement would demand compute and data volumes beyond what is currently used for language pretraining.

\paragraph{Compute efficiency for dense models.}
\label{sec:scaling_dense_comparison}

We compare Dense Multimodal with RAE (SigLIP 2) against two unimodal baselines: Text-Only and T2I-Only. To characterize scaling near the entropy floor, we fit $L(C) = A \cdot C^{-\alpha} + E$~\citep{kaplan2020scaling}, where $E$ is irreducible loss, selecting $E$ via grid search to minimize log-space MSE.
\Cref{fig:scaling_dense} shows that the unified model matches or exceeds unimodal baselines. For \Language, Dense Multimodal matches the Text-Only baseline across the entire compute range, reaching Notes PPL 23.7 vs.\ 23.8 and DCLM PPL 13.3 vs.\ 13.6 at $10^{21}$ FLOPs. For \Vision, Dense Multimodal outperforms T2I-Only (DPG 0.622 vs.\ 0.598; FID 39.3 vs.\ 41.5), suggesting that joint language training provides a complementary signal that improves generation quality.

\label{sec:scaling_moe}
\begin{figure}[t]
  \centering
    \includegraphics[width=0.98\linewidth]{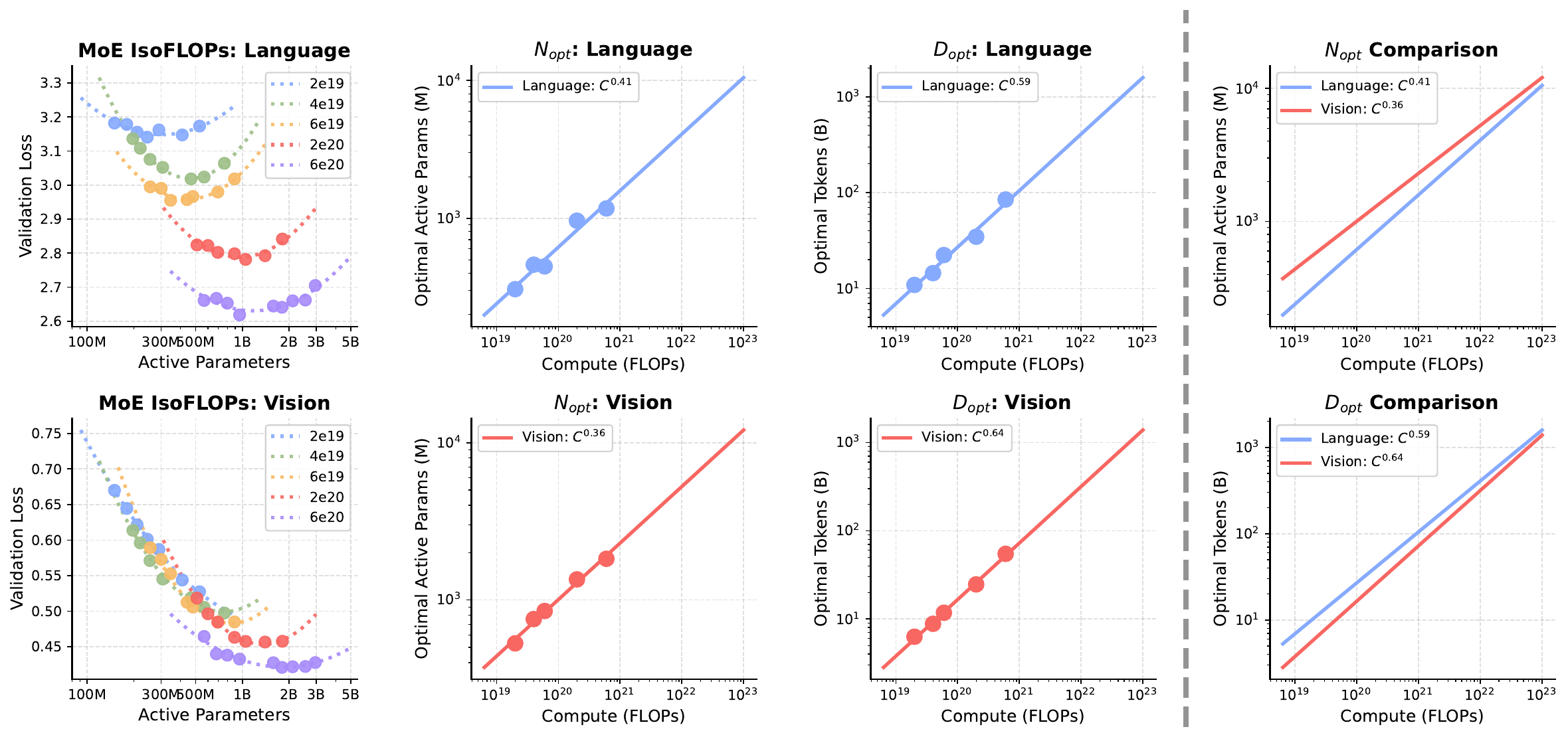}
  \caption{\textbf{Scaling laws for multimodal MoE models.} MoE narrows the vision-language exponent gap from 0.10 (dense) to 0.05 ($D_{\text{opt}} \propto C^{0.64}$ for vision vs.\ $C^{0.59}$ for language).}
  \label{fig:scaling_moe_combined}
\end{figure}

\begin{figure*}[t]
  \centering
  \includegraphics[width=0.98\linewidth]{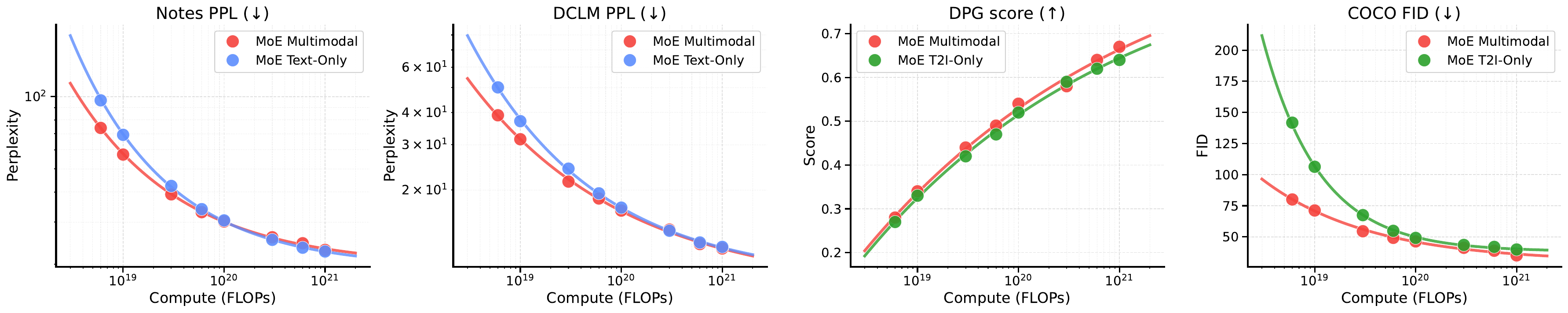}
  \vspace{-1mm}
  \caption{\textbf{Scaling laws and efficiency benchmarks for unified MoE models.} MoE Multimodal closely tracks unimodal MoE baselines across all four metrics, with near-identical FID and competitive language perplexity at $10^{21}$ FLOPs.}
  \label{fig:scaling_moe}
  \vspace{-1mm}
\end{figure*}

\paragraph{Compute-optimal for MoE models.}
We evaluate MoE scaling with a sparsity ratio of 16 (16$\times$ more total experts than activated). As established in \Cref{subsec:moe_sparsity}, both modalities benefit from higher sparsity. \Cref{fig:scaling_moe_combined} presents the MoE IsoFLOP results. Under this sparse regime, \Language exhibits $a\approx0.41$, $b\approx0.59$, while \Vision exhibits $a\approx0.36$, $b\approx0.64$. MoE halves the parameter scaling exponent gap from 0.10 (dense) to 0.05.

Moving to MoE reduces the scaling asymmetry: the language data exponent $b$ increases from 0.53 to 0.59, moving closer to vision's data-intensive regime. Because both modalities benefit from higher sparsity, further increasing the total expert count provides a mechanism to further close this gap. Sparsity thus serves as a practical architectural lever to balance the divergent data demands of unified models.

\paragraph{MoE compute efficiency.}
\Cref{fig:scaling_moe} compares MoE Multimodal with RAE (SigLIP 2) against unimodal MoE baselines across the full compute range. Across all four metrics, the unified model closely tracks its unimodal counterparts. At $10^{21}$ FLOPs, it reaches DCLM PPL 12.3 vs.\ 12.0 for MoE Text-Only, and FID 39.2 vs.\ 39.8 for MoE T2I-Only. MoE enables a single model to match unimodal performance on both modalities with minimal overhead, effectively leveraging sparse expert capacity.

\section{Related Work}
\label{sec: related work}

\paragraph{Multimodal pretraining.}
Multimodal pretraining involves training models from scratch on mixed-modal data, and is distinct from finetuning pretrained backbones. Multimodal pretraining can be divided into \textit{understanding-only} and \textit{unified understanding and generation} approaches. Understanding-only models~\citep{li2024aria,shukor2025scaling} output only text, using visual data to condition language modeling. Unified methods output both text and images: early work quantized images into discrete tokens~\citep{vqvae, vqvae2, rqvae} for next-token prediction~\citep{lu2022unified,aghajanyan2022cm3, team2024chameleon, lu2024unified, emu3}, while recent work~\citep{zhou2024transfusion,cui2025emu3} incorporates diffusion. We build on Transfusion and systematically study the design space---visual representation, architecture, and data---for unified visual understanding and generation. 

\paragraph{Unified models and their architectures.}

Unified models~\citep{zhang2025unified, xiao2025mindomni, geng2025x, xu2025tbac, xin2025lumina, li2025onecat,wang2025lightbagel,wei2025univideo,wang2025ovis, li2025uniworld,  nguyen2025oneflow,metaquery,li2025synergen, zhang2026nextflow} input and output multimodal data---typically vision and language. One line of work~\citep{dai2023emu,emu2,ge2024seed,dong2023dreamllm,tong2024metamorph, metaquery} connects pretrained language models~\citep{touvron2023llama, grattafiori2024llama3} to pretrained diffusion models~\citep{LDM, podell2023sdxl,flux,xie2025sana}, leveraging the LLM to provide conditioning for the diffusion model, typically via adapter training or end-to-end finetuning. More recent works~\citep{liang2024mixture,lmfusion,ma2025janusflow,chen2025januspro,mogao,showo,deng2025bagel,xie2025show} train diffusion models \textit{from scratch} jointly with the language model. These works incorporate different architectural designs, varying levels of separation between language and diffusion models. Transfusion~\citep{zhou2024transfusion} and the Janus series~\citep{janus, ma2025janusflow, chen2025januspro} leverage a single transformer to handle both language and vision modeling, while other recent approaches introduce stricter separation, such as splitting FFNs~\citep{lmfusion, lin2024moma} or attention blocks~\citep{liang2024mixture,deng2025bagel}. Recent studies~\citep{OpenAI2024gpt4o,tong2024metamorph,mogao} have shown that unification offers advantages over training T2I models alone, enabling models to leverage richer textual priors for visual generation---unified models outperform specialized generation-only models on benchmarks~\citep{wise, li2026ueval,yang2026ureason} like WISE. We compare these architectures systematically in~\Cref{sec:arch_ablations} and demonstrate that instead of manually designing modality separation, we can employ MoE architectures to learn the optimal degree of separation from data.

\paragraph{Visual representations.}
Understanding and generation have historically used separate representations. Understanding relies on high-dimensional semantic latents from language-supervised~\citep{radford2021learning, zhai2023sigmoid, sun2023eva, sun2024eva, chen2024internvl, xu2023demystifying,tschannen2025siglip} or self-supervised~\citep{simclr, MAE, moco,caron2021emerging, fan2023motion, Dinov2, fan2025scaling, xu2025next} models. Generation favors low-dimensional VAE latents~\citep{VAE, LDM, flux, cao2025hunyuanimage}. Recent work~\citep{chen2025vugen, chen2025aligning,zheng2025diffusion, scale-rae-2026} shows semantic latents also enable strong generation. We demonstrate that VAE latents degrade visual understanding (\Cref{fig:native_vs_finetune}), while semantic latents with RAE excel at both tasks.

\paragraph{World models.}
World models~\citep{118723, worldmodel, lecun2022path} enable test-time reasoning for embodied tasks. This paradigm evolved from in-domain models for policy training~\citep{dreamer, dreamerv2, tdmpc, transformer100k, iris,dreamerv3, tdmpc2,dreamerv4} to large-scale video generators~\citep{genie, sora2,wan2025wan,veo3}. Recent approaches~\citep{dinowm, bar2025navigation, vjepa2, dexwm, VLWM} leverage pretrained vision-only or vision-language representations for planning,  but still require in-domain trajectories. We investigate whether world modeling emerges directly from multimodal pretraining, reducing the need for domain-specific data.

\paragraph{Scaling laws.}
Language model loss scales as a power law with compute~\citep{kaplan2020scaling, radford2018improving, brown2020language}. Chinchilla~\citep{hoffmann2022chinchilla, muennighoff2023scaling} quantified compute-optimal parameter-token tradeoffs, which were later extended to MoE~\citep{tian2025towards} and diffusion models~\citep{polyak2025moviegencastmedia}. We conduct Chinchilla-style IsoFLOP analysis for joint vision-language scaling, revealing an asymmetry between modalities that MoE architectures help harmonize (\Cref{sec:scaling}).

\section{Discussion}
\label{sec:discussion}

We set out to bring empirical clarity to the design space of unified multimodal pretraining. A central question in this space is whether vision and language can coexist in a single model without mutual degradation. By training from scratch and systematically studying one variable at a time, we find that modality competition is not a fatal flaw of multimodal pretraining---it is just a symptom of specific design choices.

\paragraph{A single visual representation suffices.}  Previous work has assumed that separate visual representations are necessary for understanding and generation~\citep{deng2025bagel, janus}. We join a growing body of work~\citep{svgt2i2025,chen2025vugen,scale-rae-2026,zheng2025diffusion,shi2025latent} that challenges the necessity of this dichotomy. Our results show that a single high-dimensional semantic representation (RAE) excels at both visual understanding and generation (\Cref{sec:ablation_visual_representation}), and our MoE analysis confirms that the same experts are often activated for both tasks (\Cref{subsec:moe_analysis}). This convergence was fully learned from the data without any human priors. Furthermore, RAE representations continue to improve as capacity scales, while VAE-based methods exhibit loss saturation (\Cref{fig:moe_loss}), suggesting that semantic representations are better suited as the foundation for scaling unified models.

However, we note that current semantic vision encoders can still lag behind VAEs in fine-grained reconstruction~\citep{zhang2025both}. We hope our findings will encourage the community to close this gap by developing better generation-aware semantic representations.

\paragraph{Modality competition is largely solvable.}  A common assumption is that vision and language are inherently at odds within a single model, meaning that training on one modality necessarily degrades the other~\citep{lin2024moma}. Our findings paint a different picture: the ``modality tax'' has two identifiable sources, neither of which is the visual modality itself. The first is data: friction stems from distributional shifts in image-text captions, not from vision itself (\Cref{sec:data_composition}). Pure video is complementary to language, and general multimodal pretraining yields positive transfer for VQA and world modeling (\Cref{sec:data_synergy,sec:world_modeling}). The second is architectural: dense models rigidly allocate capacity between modalities, which modality-specific FFNs partially address, and MoE further resolves by learning to allocate capacity per token (\Cref{sec:moe}). We note that multimodal pretraining can slightly degrade out-of-distribution text generalization, but the in-distribution impact is minimal (\Cref{fig:data_ablation_text}), and expect this gap to narrow with more training. Rather than viewing vision as a competitor, we can leverage vast quantities of visual data, such as pure video, as a strategic resource for the next generation of models.

\paragraph{Efficient and effective scaling with MoE.}  MoE has proven effective for scaling language models~\citep{deepseekv2, deepseekv3}, but making it work for multimodal pretraining requires additional design. We find that high granularity is essential and that the choice of diffusion prediction target depends on the visual representation (\Cref{sec:moe}). With these design choices, both modalities consistently improve as total experts grow at fixed active compute (\Cref{fig:moe_sparsity}), confirming that sparse MoE scaling extends to the multimodal setting.

Our IsoFLOP analysis (\Cref{sec:scaling}) reveals that MoE resolves a fundamental scaling asymmetry: in dense models, language follows Chinchilla-like balanced allocation while vision is significantly more data-hungry, making it impossible to optimize both simultaneously. However, in the sparse MoE regime, language scaling shifts toward a more data-hungry regime, aligning with vision scaling. This suggests that sparsity plays a deeper role in multimodal models than efficiency alone---it provides the structural
flexibility for enabling modalities with fundamentally different scaling behaviors to coexist.

\paragraph{What's next?}  
The finding that world modeling capabilities emerge from general multimodal pretraining with minimal domain-specific data (\Cref{sec:world_modeling}), aligns with the vision that general-purpose models, trained broadly on how the world looks and moves, can develop an internal model of physical reality~\citep{lecun2022path}. The ingredients are already available; vast quantities of unlabeled video remain largely untapped, and our results show they can be incorporated without degrading language capabilities. As unified models grow in capacity and data, we believe the boundary between multimodal models and world models will blur, and future models will evolve into truly native, end-to-end multimodal systems. This work offers a clearer understanding of the core challenges in unified multimodal pretraining, and we hope that our insights will encourage the community to extend the foundation model paradigm beyond just LLMs, towards systems that can understand and reason about the physical world~\citep{lecun2022path}.

\section*{Acknowledgements}
We thank Jimmy Yang, Sharut Gupta, Jiachen Zhu, Daniel Bolya, Nanye Ma, Muzi Tao, Ji Lin, Tianhong Li, Ang Li, Jiawei Zhao, Jihan Yang, Shusheng Yang, Aniket Didolkar, Druv Pai, Weijia Shi, Xin Wen, Hu Xu, Xiaochuang Han, Emily Dinan, Tushar Nagarajan, Jiawei Yang, Ziqiao Ma for their helpful suggestions.

\newpage

\phantomsection
\addcontentsline{toc}{section}{References}

\bibliography{paper}
\bibliographystyle{arxiv-numbered}

\clearpage
\appendix

\addtocontents{toc}{\protect\setcounter{tocdepth}{1}}

\phantomsection
\addcontentsline{toc}{section}{Appendix}

\section*{Appendix}\label{app:appendix}

This appendix provides supplementary analyses and implementation details supporting the main paper:
\begin{itemize}[noitemsep,topsep=0pt,parsep=0pt,partopsep=0pt,leftmargin=1.5em]
    \item \S\ref{sec:limitations} discusses limitations of our work and directions for future research.
    \item \S\ref{sec:extended_methodology} provides detailed explanations of unified multimodal pretraining mechanics, including sequence formatting, hybrid attention masking, joint flow matching, and the modal-switching inference algorithm.
    \item \S\ref{sec:ablation_ffn_shared_vs_specific} compares shared vs.\ modality-specific FFN architectures for cross-modal learning.
    \item \S\ref{sec:additional_experiments} showcases more explorations: model properties in different layers, an adaptive loss centering mechanism, and whether vision is complementary to language modeling.
    \item \S\ref{sec:implementation_details} provides architecture, pretraining, inference, vision encoder, data composition, VQA evaluation, and MoE implementation details.
    \item
    \S\ref{appendix:app_wm} provides world modeling implementation and evaluation details, and more qualitative examples of zero-shot rollouts and counterfactual simulation with free-form natural language actions.
    \item \S\ref{appendix: data} describes training data sources, recaptioning procedures, video action annotations, and additional data analyses.
\end{itemize}

\section{Limitations and Future Work}
\label{sec:limitations}
We focus on investigating multimodal pretraining from scratch rather than post-training. Unified multimodal models can improve greatly through RL or even multimodal RL by enabling the system to both generate and interpret visual latents. On the architecture front, while we explore many facets of multimodal Mixture-of-Experts (MoE) design, our scope is not exhaustive; for instance, the uneven distribution of tokens per expert remains a significant bottleneck for hardware utilization. Conversely, leveraging higher granularity may enhance the scaling efficiency of these MoEs. Regarding data, our work excludes interleaved data, which remains a critical area for future exploration. In the long run, we believe the capability to spontaneously generate visual latents will drive the convergence of multimodal models and world modeling, fostering "System 2" behaviors that are grounded in the real world.

\section{Extended Methodology: Unified Multimodal Mechanics}
\label{sec:extended_methodology}

We provide more detailed explanations of our unified pretraining and inference mechanics. Building upon the Transfusion framework, our model processes discrete text tokens and continuous visual inputs within a single, shared transformer backbone.

\paragraph{Sequence formatting.} Text tokens are mapped to vectors via a standard \texttt{nn.Embedding} layer. Visual inputs are mapped to latent patches using a frozen visual encoder and flattened into 1D sequences. Because videos are encoded frame-by-frame, we enclose each individual image or video frame with beginning-of-image (\texttt{<BOI>}) and end-of-image (\texttt{<EOI>}) marker tokens. Consequently, a video is processed as a series of individually bounded frames within the sequence (e.g. \texttt{<BOI>} frame 1 \texttt{<EOI>} \texttt{<BOI>} frame 2 \texttt{<EOI>}).

\paragraph{Hybrid attention masking.} To process this mixed sequence, we implement a hybrid attention mask using FlexAttention. Text tokens apply a standard causal mask, attending only to preceding elements in the sequence. For visual data, we apply a block-wise causal mask: patches within the exact same image (or video frame) attend to one another bidirectionally, but can only attend causally to text or visual frames that appeared previously in the sequence. This ensures the model captures full spatial context within a single frame while preserving the strict autoregressive structure of the overall multimodal sequence.

\paragraph{Joint training and flow matching.} During training, we apply distinct objectives to each modality within the mixed sequence. For text, the input tokens remain discrete, and the model is trained using standard next-token prediction. For visual data, we employ frame-wise flow matching. Given a clean sequence of visual latent tokens $z_0$, we sample an independent timestep $t \sim \mathcal{U}[0, 1]$ for the entire image or video frame. We inject Gaussian noise to create the interpolated noisy latents $z_t$, which are then fed into the transformer alongside the text tokens. Consequently, any text tokens appearing after an image in the sequence natively condition on these noisy visual representations. The transformer then simultaneously predicts the next discrete text tokens and the continuous velocity field $v_\theta$ required to denoise $z_t$ back to $z_0$.

\paragraph{Loss computation and routing.} The model optimizes the joint objective $\mathcal{L}=\lambda_{LM}\mathcal{L}_{LM}+\lambda_{flow}\mathcal{L}_{flow}$ in a single forward-backward pass. The transformer's output hidden states are routed to modality-specific heads based on their sequence position. States corresponding to discrete text project to a vocabulary head to compute the cross-entropy loss $\mathcal{L}_{LM}$ per token. Conversely, tokens within the \texttt{<BOI>} and \texttt{<EOI>} boundaries project to a continuous linear head to predict the velocity field, computing the flow matching loss $\mathcal{L}_{flow}$ over the entire visual frame.

\paragraph{Modal-switching inference algorithm.} At inference, the model dynamically transitions between discrete autoregressive generation and continuous flow matching. In text mode, the model samples discrete tokens autoregressively. Upon sampling a \texttt{<BOI>} token, the decoding algorithm suspends text generation and appends a sequence of pure noise vectors (e.g. 256 tokens for a $16 \times 16$ grid) to the context. The model then denoises these patches using a 25-step Euler sampler, predicting the velocity field and updating the continuous tokens in place at each step. Once the visual frame is fully generated, an \texttt{<EOI>} token is appended, and the model resumes standard text generation.
\section{Shared vs.\ Modality-Specific FFNs}
\label{sec:ablation_ffn_shared_vs_specific}

Recent work demonstrates that ``modality separation'' improves cross-modal learning, but differs in the degree of architectural decoupling~\citep{lin2024moma, liang2024mixture, lmfusion, deng2025bagel}. For example, some works use modality-specific FFNs, MoE with fixed experts per modality, and/or modality-specific attention blocks. 

We study the simplest instantiation of architectural decoupling: shared self-attention with separate vision-only and language-only FFNs (``modality-specific FFN''). We compare this to a model that instead uses a shared FFN per layer for all tokens (``shared FFN''), but is otherwise identical. While modality-specific FFN increases total parameter count, training and inference FLOPs remain constant as only one FFN is activated per token. We conduct our comparison at a scale of $\approx$1T tokens (520B text + 520B multimodal) to ensure that both models are well-trained.

\Cref{fig:ffn_shared_vs_specific} demonstrates that modality-specific FFN reduces text perplexity while simultaneously improving performance in image generation and visual understanding, compared to shared FFN. Consequently, we adopt the modality-specific FFN architecture as the default setting, unless otherwise specified. In \Cref{sec:moe}, we further extend the idea of modality separation by exploring MoE architectures.

\section{Additional Results and Exploration}
\label{sec:additional_experiments}

\begin{figure}[htbp]
\centering
\includegraphics[width=\textwidth,height=\textheight,keepaspectratio]{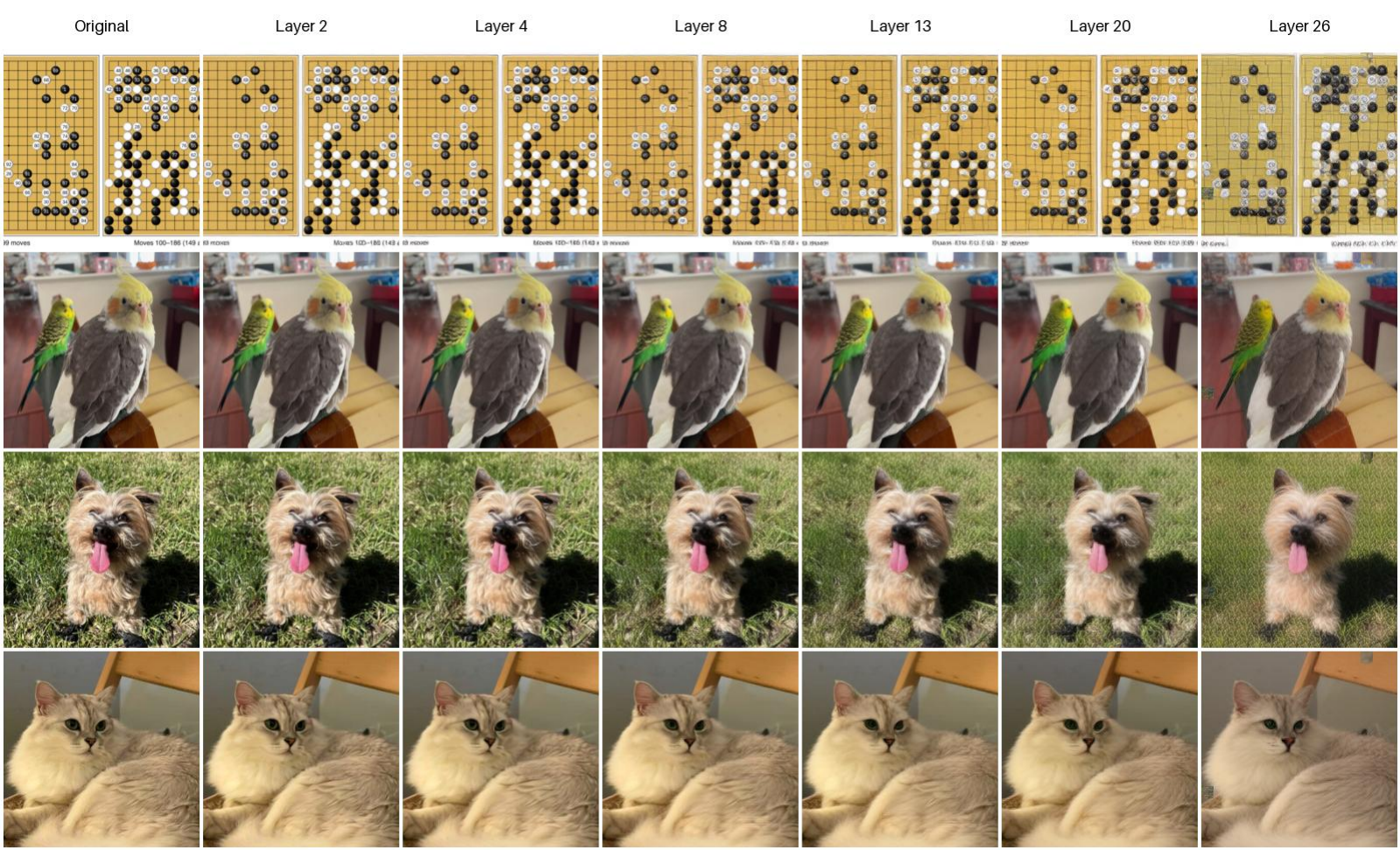}
  \caption{Qualitative reconstruction results across various SigLIP 2 layers using RAE. Observations indicate that earlier layers yield higher-fidelity reconstructions. We trained our own decoder over 3M images for this ablation.}
  \label{fig:rae_recon_qualitative}
\end{figure}

\subsection{Layer-wise Performance}
\label{sec:layerwise_perf}
\begin{figure}[h]
\centering
\begin{subfigure}[t]{0.48\linewidth}
\centering
\includegraphics[width=\linewidth]{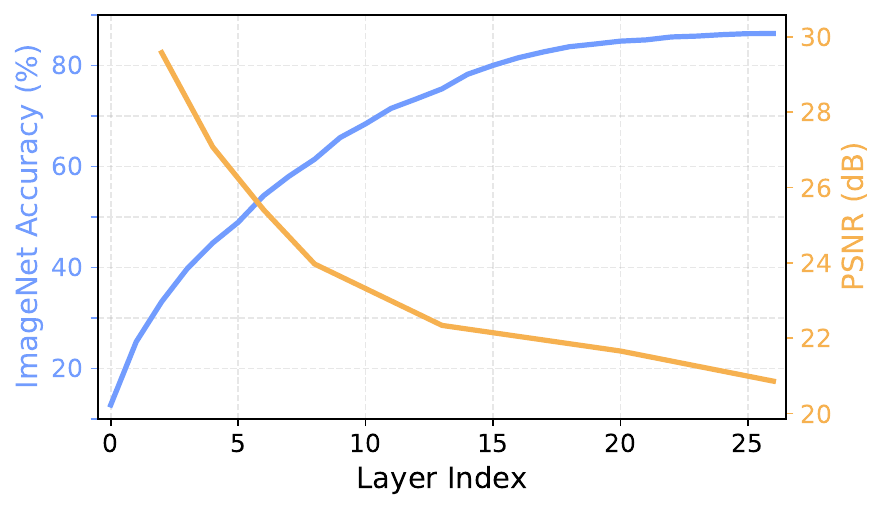}
\caption{SigLIP~2 encoder}
\label{fig:layerwise_encoder}
\end{subfigure}
\hfill
\begin{subfigure}[t]{0.48\linewidth}
\centering
\includegraphics[width=\linewidth]{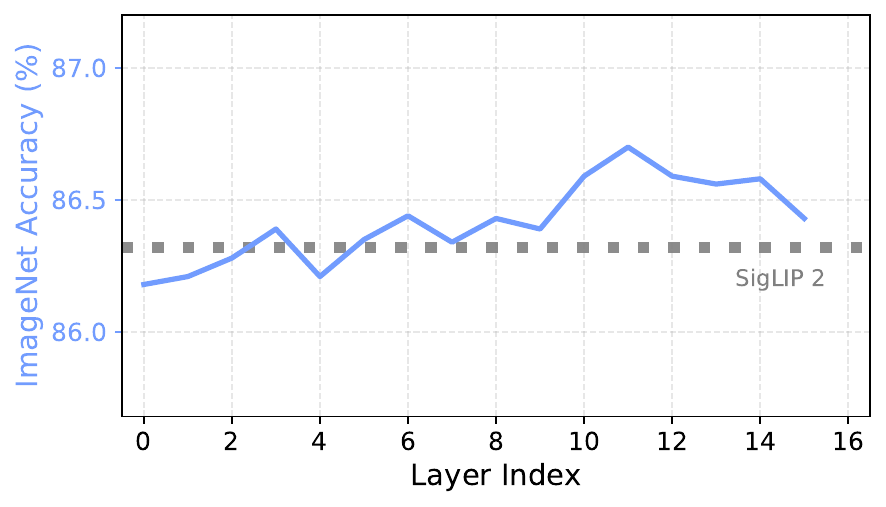}
\caption{Transformer backbone}
\label{fig:layerwise_transformer}
\end{subfigure}
\caption{Layer-wise linear probing results. \textbf{(a)}~ImageNet accuracy and PSNR across SigLIP~2 encoder layers reveal an inverse relationship between semantic understanding and pixel fidelity. \textbf{(b)}~The Transformer backbone maintains high ImageNet accuracy throughout all layers; the dotted line marks SigLIP~2's last-layer accuracy for reference (86.3\%).}
\label{fig:layerwise_accuracy}
\end{figure}

Recent work shows that visual features exhibit different properties throughout different layers of the vision encoder~\citep{bolya2025perception}. We use layer-wise linear probing to study two questions: (1)~how semantic understanding and low-level pixel fidelity evolve across layers of the SigLIP~2 encoder, and (2)~whether the transformer backbone preserves the encoder's semantic representations after training. For each layer, we train a linear classifier on frozen features and report ImageNet~\citep{deng2009imagenet} accuracy to measure semantic understanding, and PSNR to measure pixel-level reconstruction quality. Quantitative results are shown in \Cref{fig:layerwise_encoder,fig:layerwise_transformer}.

\paragraph{Representation encoder.} Across SigLIP~2 layers, ImageNet accuracy rises steadily from 12.8\% (layer~0) to 86.3\% (layer~26), while PSNR drops from 29.6\,dB to 20.9\,dB. This reveals a fundamental trade-off in which deeper encoder layers progressively discard fine-grained spatial information in favor of increasingly abstract semantic features. In practice, our models train on the last-layer features of SigLIP~2, which maximizes semantic content but leaves improvements in generation quality on the table due to the degraded pixel fidelity at that depth. Training on features from multiple layers could in principle recover this fidelity, but at the cost of simplicity. This underscores the need for future vision encoders that achieve a better balance between semantic abstraction and spatial fidelity within a single representation. As such, developing generation-aware features represents a promising direction for future research. We show qualitative examples in \Cref{fig:rae_recon_qualitative}.  Finally, we note that these observations are conditioned on the specific MetaCLIP training mixture used to train the decoder. Recent studies~\citep{scale-rae-2026} indicate that scaling the data volume and targeting specific data distributions when training the decoder, can mitigate this degradation in fidelity.

\paragraph{Transformer backbone.} We evaluate the visual representations within the network using ImageNet-1K linear probing. Despite the complexity of the joint training mixture---which spans language modeling, video prediction, text-to-image, image-to-text, and action-conditioned prediction---the learned transformer preserves the semantic quality of the features. Specifically, the linear probe accuracy of the backbone representations slightly surpasses that of the frozen SigLIP input, exhibiting minor improvements across successive layers. This indicates that high-quality visual representations are preserved within the learned model. We hypothesize that this retention is driven by two complementary training dynamics: the image-to-text (I2T) objective functions as caption-based representation learning~\citep{tschannen2024image, wan2024locca}, while the continuous diffusion objective implicitly learns rich semantic representations~\citep{li2023your, repa}.

\subsection{Loss Centering}
\label{sec:loss_centering}
Here, we study additional optimization techniques in multimodal training. In the standard Transfusion framework, the joint loss $\mathcal{L}=\lambda_{\text{LM}}\mathcal{L}_{\text{LM}}+\lambda_{\text{flow}}\mathcal{L}_{\text{flow}}$ uses fixed scalar weights throughout training (e.g. $\lambda_{\text{LM}}{=}1$, $\lambda_{\text{flow}}{=}3$). However, different visual encoders produce flow matching losses of vastly different magnitudes. When the vision loss is large relative to the text loss, a fixed weighting scheme might over-emphasize the visual objective, degrading language performance. Conversely, when the vision loss is small, the model under-invests in generation quality. A single fixed ratio cannot accommodate this variation across encoders nor across training stages as losses evolve.

\paragraph{Method.} Inspired by the centering mechanism in DINO~\citep{caron2021emerging}, we replace the fixed loss weights with adaptive, per-step weights derived from an exponential moving average (EMA) of the loss magnitudes. At each training step, we compute a weighted center of the current losses:
\begin{equation}
  c_{\text{current}} = \alpha \, \mathcal{L}_{\text{flow}} + (1-\alpha) \, \mathcal{L}_{\text{LM}},
\end{equation}
where $\alpha \in [0,1]$ controls the desired emphasis between vision and text. The running center is updated via EMA: $c_{\text{target}} \leftarrow \mu \, c_{\text{target}} + (1-\mu) \, c_{\text{current}}$, with momentum $\mu$. The per-modality weights are then:
\begin{equation}
  w_{\text{flow}} = \frac{c_{\text{target}}}{\mathcal{L}_{\text{flow}}}, \qquad
  w_{\text{LM}} = \frac{c_{\text{target}}}{\mathcal{L}_{\text{LM}}}.
\end{equation}
This normalizes each loss to the shared center, automatically up-weighting the smaller loss and down-weighting the larger one. The parameter $\alpha$ provides a tunable knob: setting $\alpha$ closer to $1$ emphasizes vision, while $\alpha$ closer to $0$ emphasizes text.

\paragraph{Results.} \Cref{tab:loss_centering} compares models trained with fixed weights (``Uncentered'') against loss centering (``Centered'') across three visual encoders. Loss centering consistently improves DPG score across all encoders, with SigLIP~2 obtaining the largest gain ($+0.042$). The trade-off is a modest increase in text perplexity, with SigLIP~2 incurring the largest penalty ($+0.30$). Notably, even after centering, SigLIP~2 retains the best DPG score (0.612) while its perplexity remains competitive. The $\alpha$ parameter offers practitioners direct control over this trade-off, enabling task-specific tuning without architecture changes.

\begin{table}[h]
\centering
\caption{Effect of loss centering on generation quality (DPG~$\uparrow$) and text perplexity (PPL~$\downarrow$) across visual encoders. Centering consistently improves DPG at a modest PPL cost.}
\label{tab:loss_centering}
\begin{small}
\begin{tabular}{@{}l cc c cc@{}}
\toprule
 & \multicolumn{2}{c}{\textbf{DPG Score}~($\uparrow$)} & & \multicolumn{2}{c}{\textbf{Perplexity}~($\downarrow$)} \\
\cmidrule{2-3} \cmidrule{5-6}
\textbf{Encoder} & Uncentered & Centered & & Uncentered & Centered \\
\midrule
SigLIP 2 & 0.570 & \textbf{0.612} \small{\textcolor{teal}{(+.042)}} & & 15.06 & 15.36 \small{\textcolor{red}{(+.30)}} \\
FLUX.1    & 0.498 & 0.510 \small{\textcolor{teal}{(+.012)}} & & \textbf{14.96} & 15.14 \small{\textcolor{red}{(+.18)}} \\
SD-VAE    & 0.470 & 0.480 \small{\textcolor{teal}{(+.010)}} & & 15.13 & 15.35 \small{\textcolor{red}{(+.22)}} \\
\bottomrule
\end{tabular}
\end{small}
\end{table}

\subsection{Text Performance on Data Composition Studies}
\label{appendix: text perf data composition}

\paragraph{Vision and language are complementary.}

\begin{figure}[h]
\centering
\includegraphics[width=0.6\linewidth]{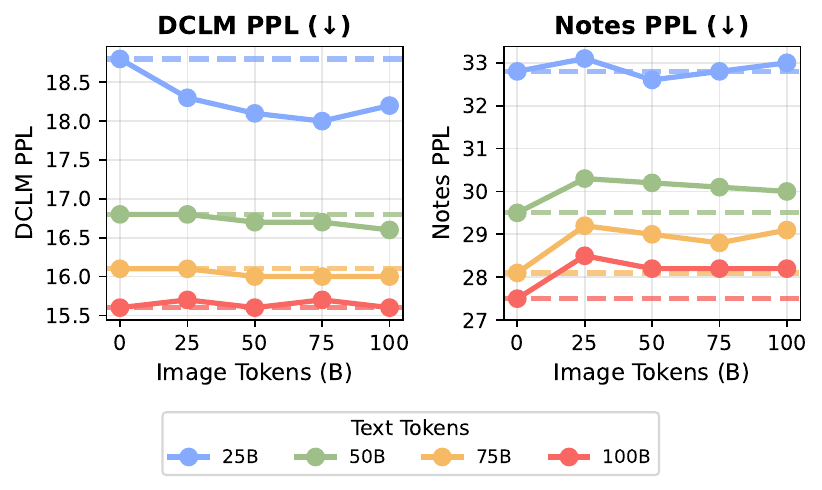}
  \caption{\textbf{Multimodal co-training exceeds unimodal performance.} We compare multimodal models (solid lines) against unimodal baselines (dashed lines) across varying budgets.}
  \label{fig:data_synergy:img_tok}
\end{figure}

We measure how each modality affects the other by training models across all combinations of $\{0, 25, 50, 75, 100\}$B text and multimodal tokens (\Cref{fig:data_synergy:img_tok}). Consistent with \Cref{sec:ablation_pretrain_data}, adding visual tokens to a fixed text budget minimally impacts text perplexity with small improvements on DCLM and slight degradation on Notes. Adding text tokens to a fixed multimodal budget consistently improves performance, clearly surpassing the vision-only baseline.

\section{Implementation Details}
\label{sec:implementation_details}

\subsection{Architecture Details}
\label{appendix:architecture_details}
Unlike Transfusion~\citep{zhou2024transfusion}, we use simple linear projection layers instead of U-Net for visual input/output projections. Our architecture is encoder-agnostic and accepts any frozen visual encoder. For the loss weighting $\mathcal{L} = \lambda_{\text{LM}}\mathcal{L}_{\text{LM}} + \lambda_{\text{flow}}\mathcal{L}_{\text{flow}}$, we use $\lambda_{\text{LM}}{=}1.0$ and $\lambda_{\text{flow}}{=}3.0$ by default (Transfusion uses $\lambda_{\text{LM}}{=}1.0$ and $\lambda_{\text{flow}}{=}6.0$).

\subsection{Pretraining Details}
\label{appendix:pretraining_details}
By default, training uses a sequence length of 4096, 128 GPUs, and a batch size of 4 per GPU, for a total batch size of $\approx$2M tokens per step. The total number of training steps is calculated according to the total training budget. We use AdamW optimizer with a peak LR of $3{\times}10^{-4}$, 1000 warmup steps, and cosine decay to 5\% of the peak LR.

\subsection{Inference Details}
\label{appendix:inference_details}

Text generation uses autoregressive next-token prediction. Visual generation denoises via a 25-step Euler sampler with classifier-free guidance scale of 3.0 and  10\% conditioning dropout during training~\citep{ho2022classifier}. For VAE encoders, we use pretrained decoders from SD-VAE~\citep{LDM} and FLUX.1~\citep{flux}. For SigLIP 2 So400m, DINOv2-L, and WebSSL-L, we use the RAE decoder~\citep{zheng2025diffusion}.

\subsection{Vision Encoders}
\label{appendix: vision_encoders}

\begin{table}[t]
\centering
\caption{Vision encoder configurations. All encoders produce 256 tokens ($16{\times}16$ grid) per image. For VAE encoders that natively produce $32{\times}32 = 1024$ latent patches, we apply PixelUnshuffle($2$)~\citep{pixelshuffle} to rearrange the spatial grid into channels, reducing to $16{\times}16 = 256$ tokens while increasing the channel dimension by $4{\times}$. To apply the diffusion loss, we then apply PixelShuffle($2$) to upsample the feature map back to the original spatial resolution.}
\label{tab:vision_encoders}
\begin{tabular}{llccccc}
\toprule
\textbf{Type} & \textbf{Encoder} & \textbf{Resolution} & \textbf{Patch Size} & \textbf{Grid Size} & \textbf{Spatial Adaptation} & \textbf{Tokens} \\
\midrule
Raw Pixel & Raw Pixel & $224^2$ & 14$\times$14 & $16{\times}16$ & --- & 256 \\
\midrule
\multirow{2}{*}{RAE}
  & SigLIP 2 So400m & $224^2$ & 14$\times$14 & $16{\times}16$ & --- & 256 \\
  & DINOv2-L   & $224^2$ & 14$\times$14 & $16{\times}16$ & --- & 256 \\
  & WebSSL-L   & $224^2$ & 14$\times$14 & $16{\times}16$ & --- & 256 \\
\midrule
\multirow{2}{*}{VAE}
  & SD-VAE     & $256^2$ & $8{\times}$8 & $32{\times}32$ & Pixel[Un]shuffle(2) & 256 \\
  & FLUX.1 VAE & $256^2$ & $8{\times}$8 & $32{\times}32$ & Pixel[Un]shuffle(2) & 256 \\
\bottomrule
\end{tabular}
\end{table}

The model receives 256 tokens per image for all vision encoder types, as summarized in \Cref{tab:vision_encoders}. Raw pixel and RAE encoders (SigLIP 2, DINOv2, WebSSL) natively produce $16{\times}16 = 256$ tokens using patch size 14$\times$14 on $224{\times}224$ images. For VAE encoders (SD-VAE, FLUX.1 VAE) that natively produce $32{\times}32 = 1024$ latent patches, we apply PixelUnshuffle($2$)~\citep{pixelshuffle} to downsample the spatial grid by increasing channel dimension, reducing to $16{\times}16 = 256$ tokens while increasing the channel dimension by $4{\times}$. This keeps the total number of vision tokens consistent for fair comparison across vision encoders. To apply the flow matching loss, we then use PixelShuffle($2$) to upsample the output of the Transformer back to the original resolution of the vision encoder output. We opt for simplicity instead of using U-Net which would introduce additional parameters.

\subsection{Data Composition}
\label{appendix:data_composition}

To study the effect of data composition on multimodal model performance, we train a series of models varying the amount of text and image tokens seen during training. Specifically, we use a 5×5 grid of data compositions: text tokens from DCLM at {0, 25, 50, 75, 100} billion and image tokens from SSTK at {0, 25, 50, 75, 100} billion, yielding 25 configurations (excluding the 0-0 case). For each configuration, we evaluate language capability using perplexity on two held-out text benchmarks (DCLM validation and Notes), and vision capability using diffusion loss and GenEval. To isolate the effect of co-training, we compare mixed-modality runs against single-modality baselines: text-only models (SSTK=0) and image-only models (DCLM=0).

\subsection{VQA Evaluation}
\label{appendix:vqa_details}
The encoder remains frozen. We finetune the multimodal model for 1 epoch on Cambrian-7M with a similar recipe as~\Cref{appendix:pretraining_details}, except with 64 GPUs (global BS of $\approx$1M tokens), and a peak LR of $1{\times}10^{-5}$. For text-only models, we first warm up the vision adapter on the Cambrian-Alignment dataset (since no encoder exists during pretraining) and keep all other parameters frozen, using a peak LR of $2{\times}10^{-3}$. We then finetune using the same recipe as the multimodal model.

We follow previous work~\citep{tong2024cambrian, fan2025scaling, han2025learning} and evaluate the finetuned model on the 16 benchmarks grouped in \citet{tong2024cambrian}: MMBench~\citep{liu2023mmbench}, Seed~\citep{ge2023planting}, MME~\citep{fu2023mme}, GQA~\citep{hudson2019gqa}, AI2D~\citep{hiippala2021ai2d}, MathVista~\citep{lu2023mathvista}, MMMU~\citep{yue2023mmmu}, ScienceQA~\citep{lu2022learn}, OCRBench~\citep{liu2024ocrbench}, TextVQA~\citep{singh2019towards}, ChartQA~\citep{masry2022chartqa}, DocVQA~\citep{mathew2021docvqa}, MMVP~\citep{tong2024eyes}, CV-Bench 2D~\citep{tong2024cambrian}, CV-Bench 3D~\citep{tong2024cambrian}, and RealWorldQA~\citep{grok}.
While we acknowledge that the scope~\citep{tao2024what, yue2024mmmu, brown2025sims,yang2025thinking,brown2025benchmark, zhou2026worldvqa, tao2026asymmetricidiosyncrasiesmultimodalmodels,yang2025cambrian} of visual understanding extends beyond these benchmarks, our purpose is to use VQA as an evaluation protocol, rather than to contend with other models as a general-purpose VQA model; we therefore leave more evaluations for future work.

\subsection{MoE models}
\label{appendix:moe_models}

\paragraph{Sparsity studies.}
To verify the efficacy of sparse scaling, we conduct experiments using a fixed computational budget per forward pass. All models are trained for approximately 57B tokens (27,248 steps) using a global batch size of 512 sequences (length 4096). Training is distributed across 64 NVIDIA H200 GPUs (8 nodes).

The base Transformer backbone consists of 16 layers with a dimension $d_{\text{model}}=2048$. We use Grouped Query Attention (GQA) with 32 query heads and 8 key-value heads. The FFN expansion multiplier is set to 1.5.

For the MoE layers, we enforce an IsoFLOP constraint by fixing the number of active parameters. We utilize a fine-grained expert dimension $d_{\text{expert}}=512$ and a router top-$k=16$. We then systematically scale the total capacity by increasing the total number of experts $E$ from 32 up to 1008. \Cref{tab:sparsity_params} details the resulting parameter counts, demonstrating that active parameters remain constant ($\approx$1.5B), while total capacity scales with the number of experts.

\begin{table}[h]
\centering
\caption{Hyperparameters and parameter counts for the sparsity (IsoFLOP) sweep. We scale total parameters by increasing the expert pool $E$ while keeping active computation constant. The slight increase in active params is due to the router. }
\label{tab:sparsity_params}
\begin{small}
\begin{tabular}{@{}lrr@{}}
\toprule
\textbf{Configuration} & \textbf{Total Params} & \textbf{Active Params} \\ \midrule
Dense Baseline & 1.50B & 1.50B \\ \midrule
$E=32$ & 2.30B & 1.50B \\
$E=64$ & 3.92B & 1.50B \\
$E=128$ & 7.14B & 1.50B \\
$E=256$ & 13.59B & 1.51B \\
$E=512$ & 26.48B & 1.52B \\
$E=1008$ & 51.46B & 1.53B \\ \bottomrule
\end{tabular}
\end{small}
\end{table}

\subsection{Analysis on MoE models}
\label{appendix:moe_analysis}
To understand how the MoE model allocates capacity across modalities, we quantify the routing preference of each expert. We perform this analysis on the $G=16$ model (256 experts, 1 modality-specific expert) trained with SigLIP 2 and $x$-prediction. Our training comprises 50\% text data, 25\% video data, 22.5\% of image-text data, and 2.5\% action data. We train for 1T tokens. 

\paragraph{Data and Routing.}
We utilize a held-out validation set consisting of text tokens from DCLM and image tokens from CC12M. We pass this data through the frozen trained model and record the selection frequency of every expert. Let $N_{\text{text}}$ and $N_{\text{image}}$ denote the total number of tokens processed for each modality. For a given expert $e_i$, let $C_{\text{text}}^{(i)}$ and $C_{\text{image}}^{(i)}$ represent the number of times that expert was selected by the router (activated in the top-$k$ set).

\paragraph{Specialization Score.}
To normalize against imbalances in the validation data size or routing budget, we compute the \emph{selection rate} $R$ for expert $e_i$ and modality $m \in \{\text{text}, \text{image}\}$ as:
\begin{equation}
  R_{m}^{(i)} = \frac{C_{m}^{(i)}}{N_{m} \times k}  
\end{equation}
where $k$ is the number of active experts per token (in our analysis, $k=16$). This represents the probability that expert $e_i$ is selected for a given token of modality $m$.

We then define the \textbf{Specialization Score} $S_i$ as the normalized difference between the modality-specific selection rates:
\begin{equation}
    S_i = \frac{R_{\text{text}}^{(i)} - R_{\text{image}}^{(i)}}{R_{\text{text}}^{(i)} + R_{\text{image}}^{(i)}}.
\end{equation}
The score $S_i$ ranges from $-1$ to $+1$. A score of $+1$ indicates the expert is exclusively selected by text tokens, while $-1$ indicates the expert is exclusively selected by image tokens.

\paragraph{Classification Categories.}
Based on the specialization score $S_i$, we classify each expert into one of three categories:
\begin{itemize}
    \item \textbf{Text Expert ($S_i > 0.5$):} The expert is significantly biased toward text. A score greater than 0.5 implies $R_{\text{text}} > 3 R_{\text{image}}$.
    \item \textbf{Vision Expert ($S_i < -0.5$):} The expert is significantly biased toward vision. A score less than -0.5 implies $R_{\text{image}} > 3 R_{\text{text}}$.
    \item \textbf{Multimodal Expert ($-0.5 \le S_i \le 0.5$):} The expert processes a balanced mix of both modalities, with neither modality exceeding the selection rate of the other by a factor of 3.
\end{itemize}

\paragraph{Visual understanding and generation experts.}

To examine whether vision experts specialize for image generation versus image understanding, we compare expert selection patterns between the two modalities using our MoE model with RAE (SigLIP 2).
For image generation (text-to-image), we process images through the diffusion denoising pipeline conditioned on text prompts, recording expert selections for image tokens.
For image understanding (image-to-text), we process images through the vision encoder and record expert selections during caption generation.
For each expert in each layer, we compute the selection rate as the fraction of tokens from each task that are routed to that expert.
We then measure the Pearson correlation between generation selection rates and understanding selection rates across all experts within each layer.

We observe strong positive correlations between expert usage for image generation and understanding across all layers. Scatter plots of expert rates show points clustered along the diagonal, indicating that experts selected frequently for generation are also selected frequently for understanding.
This suggests that the MoE routing learns shared ``vision experts'' that process visual information regardless of whether the task is generative or discriminative.

\section{World Modeling}
\label{appendix:app_wm}

\subsection{Implementation Details}
\label{appendix:world_model}

\paragraph{CEM planning.} 
We implement the Cross-Entropy Method (CEM) following Navigation World Models (NWM)~\citep{bar2025navigation}. The planner uses an 8-step horizon (2 seconds), samples $N{=}120$ candidate action sequences from a Gaussian distribution, and predicts 8 future frames per sequence. Trajectories are scored via LPIPS distance~\citep{lpips} between the final predicted frame and the goal image. We select the top $K{=}5$ trajectories and average their actions for the final plan. Each evaluation instance repeats this process 3 times; we report Absolute Trajectory Error (ATE) and Relative Pose Error (RPE) against the ground truth trajectories, following NWM.

\paragraph{Data ratio studies.}
We vary the proportion of action-conditioned navigation data in the training mixture. We use our default architecture with modality-specific FFN and SigLIP 2, trained for 200B tokens on 128 H100 GPUs.

The base mixture allocates 25\% text (DCLM), 25\% video (SSv2, Kinetics, HowTo100M, YT-Temporal), and 50\% multimodal. We ablate NWM data from 0.1\% to 25\% of total budget, with remainder split between MetaCLIP and text-action annotations. Navigation evaluation uses the RECON dataset with the CEM planner described above.

\paragraph{WASD control definition.}
We map standard keyboard controls to the robot-centric coordinate system used in NWM training (where $+x$ points forward, $+y$ points left, and yaw is measured counter-clockwise). Each key triggers a normalized displacement tuple $(\Delta x, \Delta y, \Delta\text{yaw})$. We define \textbf{Forward} (W) and \textbf{Backward} (S) as pure translations of $(+0.5, 0, 0)$ and $(-0.5, 0, 0)$, respectively. To simulate realistic motion while turning, the lateral controls include a small forward velocity component: \textbf{Turn Left} (A) maps to $(+0.2, 0, +0.5)$ and \textbf{Turn Right} (D) maps to $(+0.2, 0, -0.5)$.

\paragraph{Free-form natural language actions.}
To complement the visualizations with keyboard commands, we also demonstrate the zero-shot ability to use free-form natural language as action conditioning for navigation. For example, ``go on the road'' or 'take big steps forward''. This enables the usage of a nearly infinite set of actions, going beyond mapping to a pre-defined set of actions, which is a limitation of recent works such as Genie 3~\citep{genie, genie2, genie3}.

\subsection{More Navigation Examples}
\label{appendix:more_navigation_examples}
\begin{figure}[h]
\centering
\includegraphics[width=0.75\linewidth]{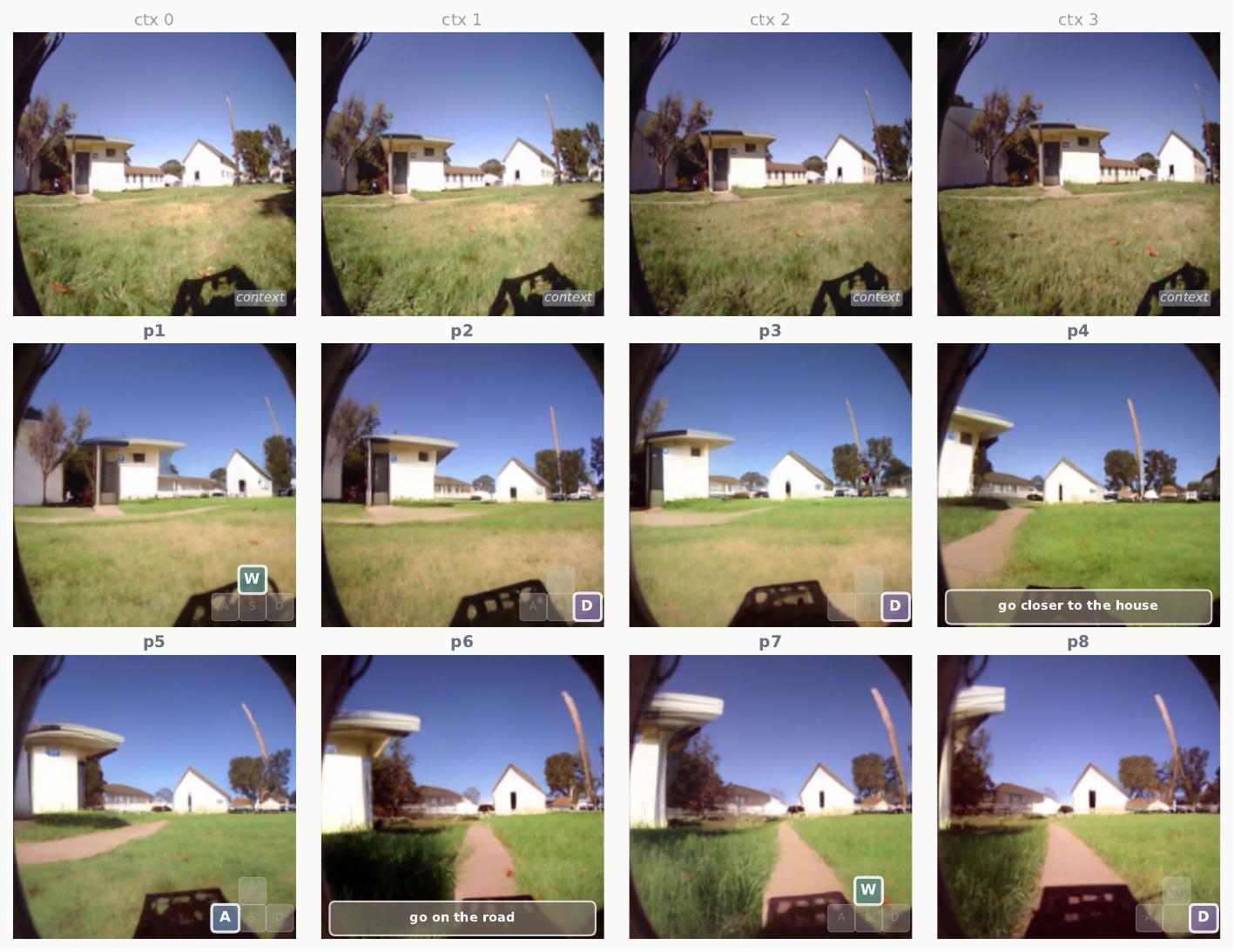}
\caption{
    \textbf{More navigation examples with natural language commands (I).}
  }\label{fig:more_nwm_qualitative-1}
\end{figure}

\begin{figure}[h]
\centering
\includegraphics[width=0.75\linewidth]{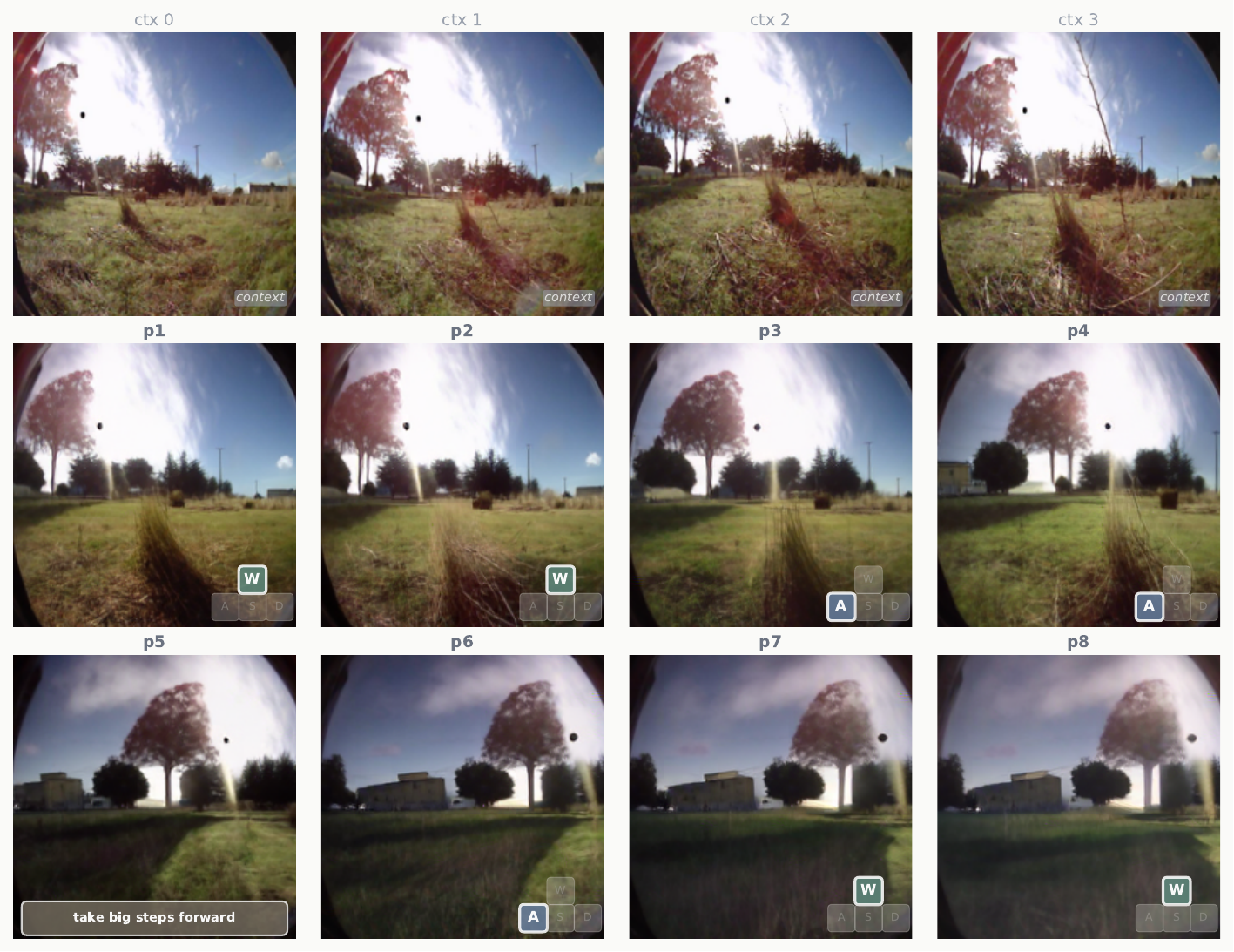}
\caption{
    \textbf{More navigation examples with natural language commands (II).}
  }\label{fig:more_nwm_qualitative-2}
\end{figure}

\begin{figure}[h]
\centering
\includegraphics[width=0.75\linewidth]{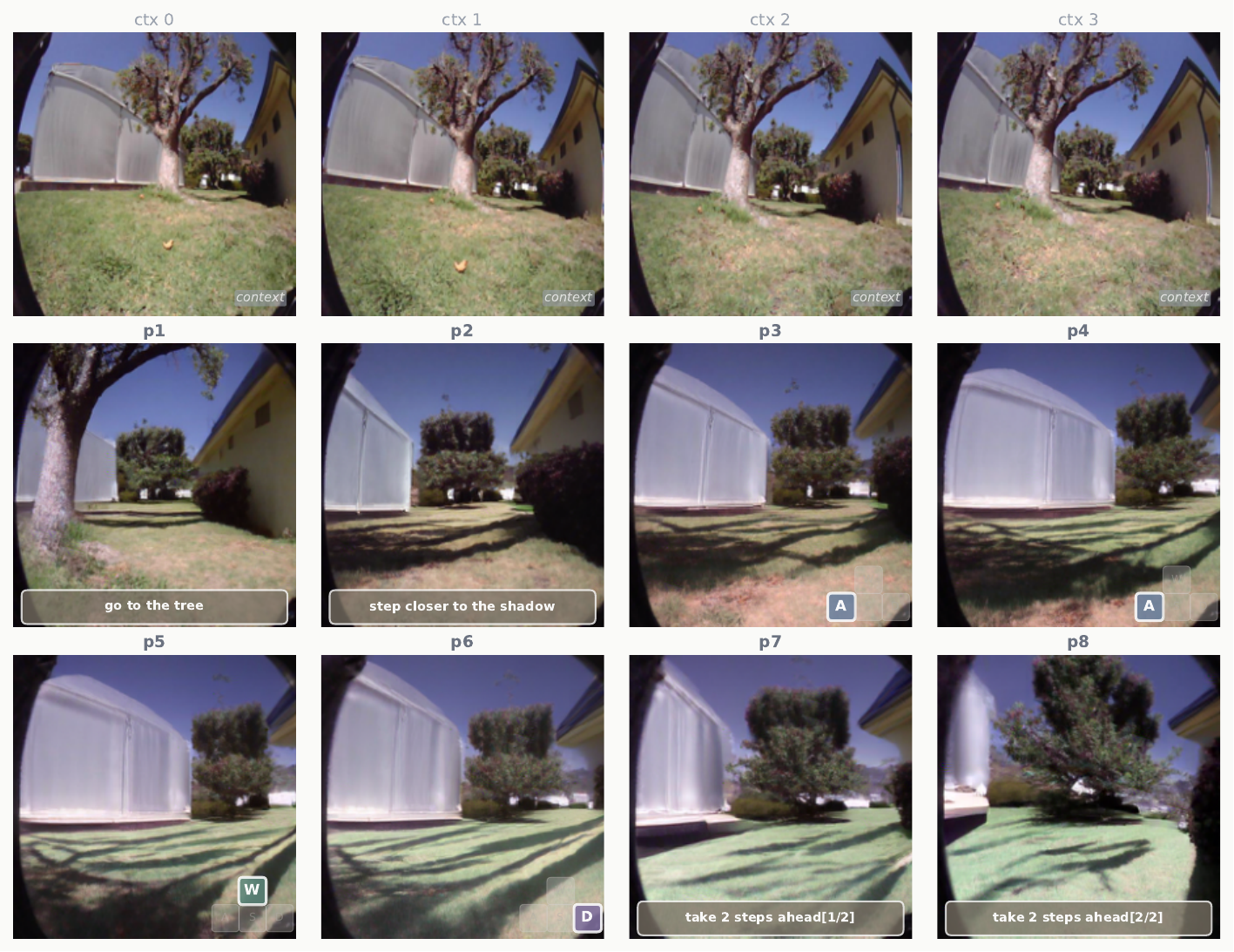}
\caption{
    \textbf{More navigation examples with natural language commands (III).}
  }\label{fig:more_nwm_qualitative-3}
\end{figure}

\begin{figure}[h]
\centering
\includegraphics[width=0.75\linewidth]{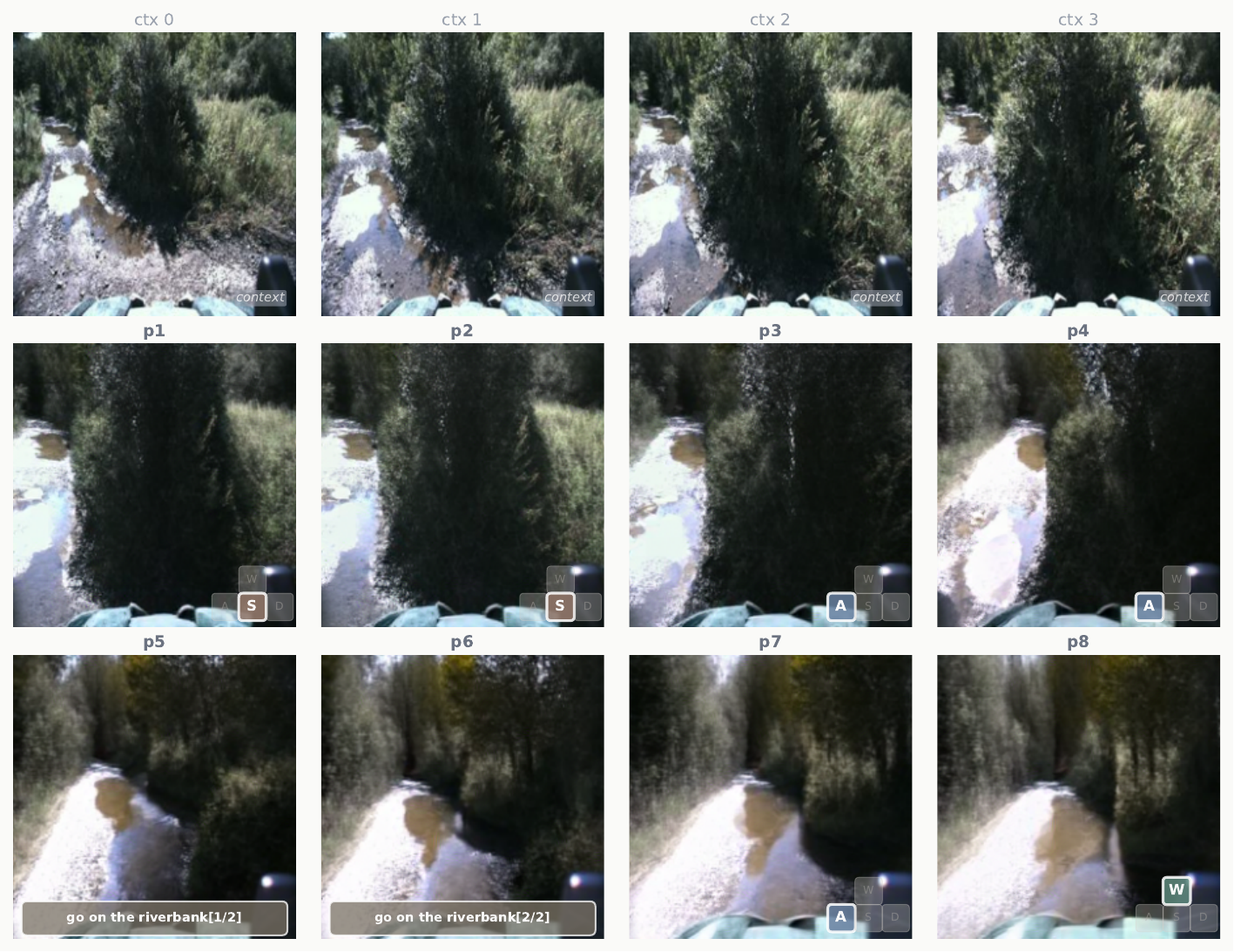}
\caption{
    \textbf{More navigation examples with natural language commands (IV).}
  }\label{fig:more_nwm_qualitative-4}
\end{figure}

\begin{figure}[h]
\centering
\includegraphics[width=0.75\linewidth]{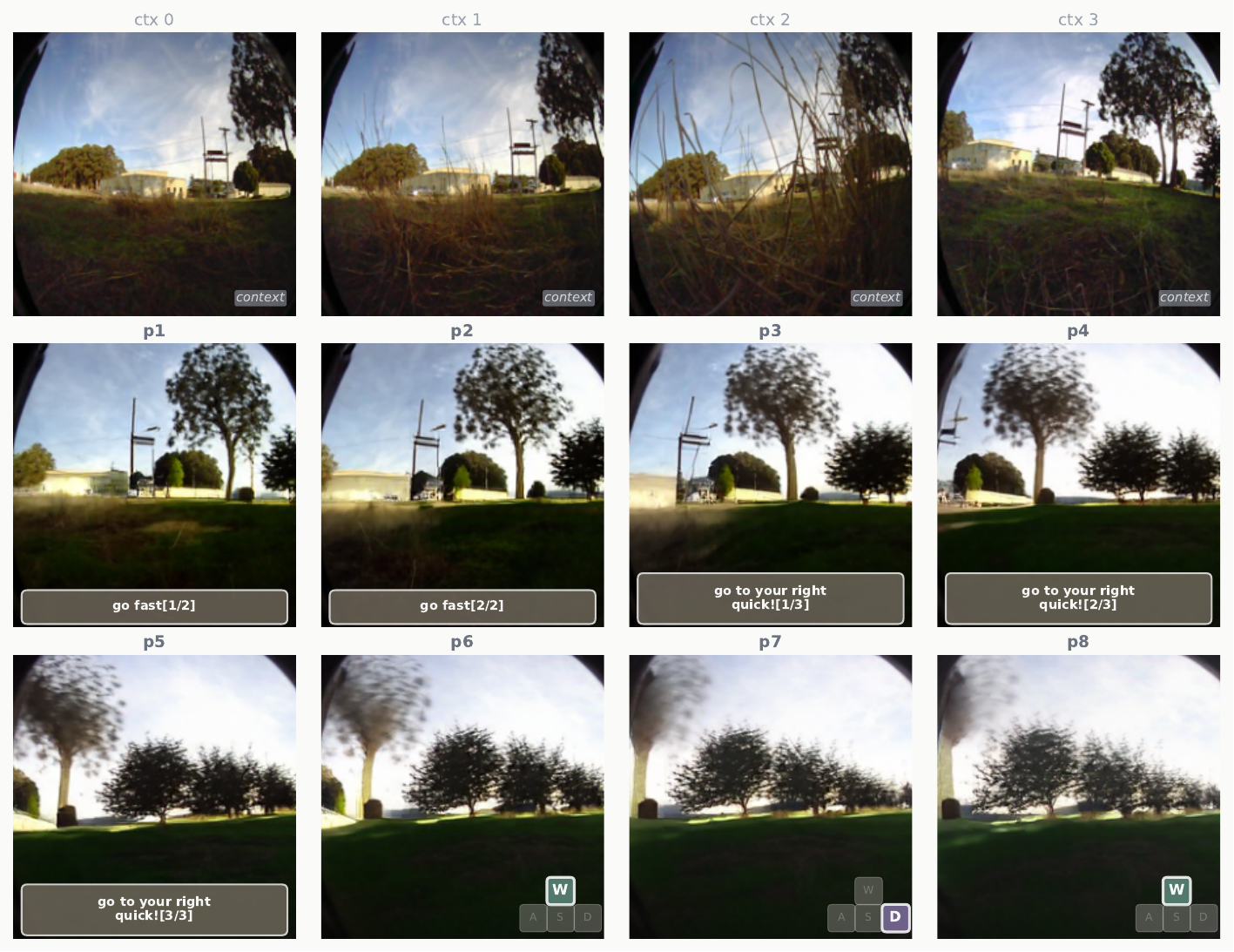}
\caption{
    \textbf{More navigation examples with natural language commands (V).}
  }\label{fig:more_nwm_qualitative-5}
\end{figure}

\begin{figure}[h]
\centering
\includegraphics[width=0.99\linewidth]{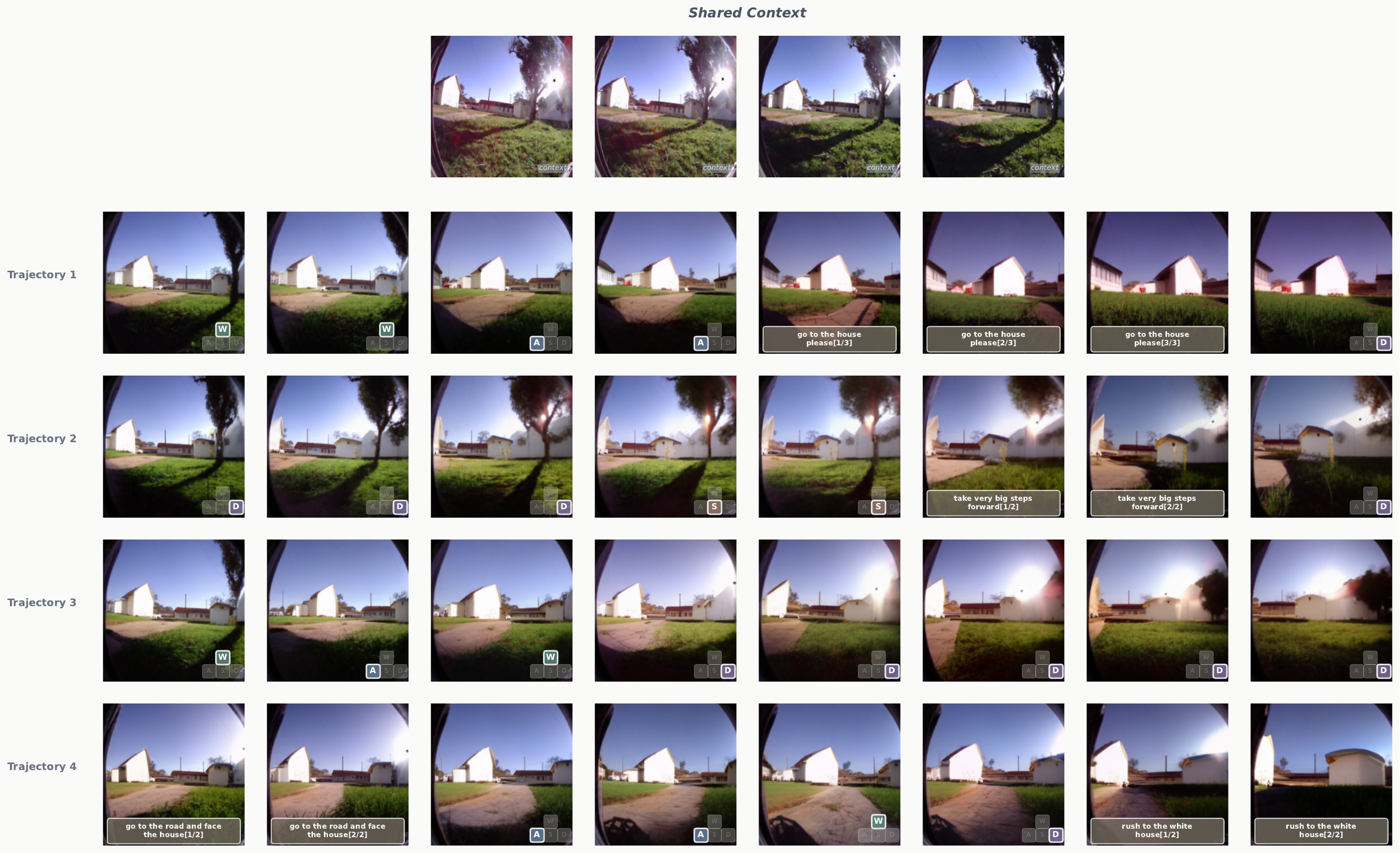}
\caption{
    \textbf{Counterfactual navigation trajectories from the same context.} The model is able to generate counterfactual trajectories following different action sequences. 
  }\label{fig:different_trajectories}
\end{figure}

We showcase more navigation rollouts with either WASD controls (Forward {\small\fbox{\texttt{W}}}, Left {\small\fbox{\texttt{A}}}, Backward {\small\fbox{\texttt{S}}}, Right {\small\fbox{\texttt{D}}}) or free-form natural language as actions, in \Cref{fig:more_nwm_qualitative-1}, \Cref{fig:more_nwm_qualitative-2}, \Cref{fig:more_nwm_qualitative-3},  \Cref{fig:more_nwm_qualitative-4}, and 
\Cref{fig:more_nwm_qualitative-5}. We also include examples in \Cref{fig:different_trajectories} of counterfactual trajectories generated from the same context but with different commands. 

\section{Data}
\label{appendix: data}

\subsection{Data Sources}
\label{appendix: data sources}
We present the detailed data sources in \Cref{tab:data_sources}.

\begin{table*}[t]
\centering
\small
\begin{tabular}{@{}ll p{8.5cm} @{}} 
\toprule
Data Type & Modality & Source \\ \midrule
Text & $\texttok \rightarrow \texttok$ & DCLM~\citep{li2024datacomp} \\ \addlinespace[2pt]
Video & $\imtok \rightarrow \imtok$ & YouTube-Temporal 1B~\citep{zellers2022merlot}, SomethingSomething V2~\citep{goyal2017something}, Kinetics~\citep{kay2017kinetics} \\ \addlinespace[2pt]
Image-Text Pairs & $\imtok\rightarrow\texttok, \texttok\rightarrow\imtok$ & MetaCLIP~\citep{xu2023demystifying}, in-house Shutterstock \\ \addlinespace[2pt]
Action-Conditioned & $\imtok + \texttok \rightarrow \imtok$ & NWM~\citep{bar2025navigation}, in-house Text-Annotated Videos \\ \bottomrule
\end{tabular}
\caption{Training data sources and supervision signals.}
\label{tab:data_sources}
\end{table*}

\subsection{MetaCLIP Recaption}
\label{appendix: metaclip recaption}

To improve caption quality, we recaption images using Qwen2.5-VL-32B-Instruct~\citep{bai2025qwen2}. We deploy the model via vLLM~\citep{kwon2023efficient} with 8-way tensor parallelism per node, enabling efficient batch inference at scale. We use nucleus sampling with temperature 0.6 and top-p 0.95, generating up to 128 tokens per caption. The system prompt used for recaptioning is shown below:

\newtcolorbox{promptbox}[1][]{
    colback=gray!5,
    colframe=gray!50,
    fonttitle=\bfseries\small,
    title=#1,
    boxrule=0.5pt,
    arc=2pt,
    left=6pt,
    right=6pt,
    top=4pt,
    bottom=4pt,
    fontupper=\small,  
}

\begin{promptbox}[Recaptioning Prompt]
\small\ttfamily
You are an expert image captioner. Write one specific, information-dense caption (1--2 sentences). Do not begin with `A', `An', or `The'. Name the main subject first, then key objects, attributes, actions, and scene context. Avoid hedging and meta phrases.
\end{promptbox}

\subsection{Video Action Annotations}
\label{appendix:video_action_annotation}
We collected about 12M video + text action annotations from YouTube-Temporal 1B~\citep{zellers2022merlot} using a model-based pipeline, where a MLLM is asked to caption the transition between two consecutive video chunks, with a focus on describing physical actions rather than just changes in appearance. The resulting annotations are then verified via self-reflection where the model is asked both whether the caption describes an action, and whether the action is reflective of the video.

For each sequence, 1-4 context frames are provided, along with a text action, and 1-4 resulting frames. Varying the number of context and resulting frames helps prevent model collapse to trivial solutions and improves robustness to video length. We leverage this data to supplement the action trajectories from NWM.

\subsection{Cosine Distance Computation}
\label{appendix:cosine_distance}

To measure the distributional similarity between DCLM and image-text datasets, we sample 100K samples from each dataset. Because some samples are long, we truncate at 2000 characters. We then extract a text embedding per sample using Qwen3-Embedding-8B~\citep{zhang2025qwen3}. Finally, we compute the cosine distance between the centroid of each dataset.

\end{document}